\documentclass[letterpaper]{article} 
\usepackage{aaai25}  
\usepackage{times}  
\usepackage{helvet}  
\usepackage{courier}  
\usepackage[hyphens]{url}  
\usepackage{graphicx} 
\usepackage{natbib}  
\usepackage{caption} 
\usepackage{algorithm}
\usepackage{algorithmic}

\usepackage{newfloat}
\usepackage{listings}
\DeclareCaptionStyle{ruled}{labelfont=normalfont,labelsep=colon,strut=off} 
\floatstyle{ruled}
\newfloat{listing}{tb}{lst}{}
\floatname{listing}{Listing}
\usepackage{amsfonts}
\usepackage{amsmath}
\usepackage{adjustbox}
\usepackage{multirow}
\usepackage{diagbox}
\usepackage{subcaption}
\usepackage{xcolor}

\nocopyright 

\title{ProGen: Revisiting Probabilistic Spatial-Temporal Time Series Forecasting from a Continuous Generative Perspective Using Stochastic Differential Equations}
\author{
    Mingze Gong,
    Lei Chen,
    Jia Li
}
\affiliations{
    \textsuperscript{\rm }Hong Kong University of Science and Technology (Guangzhou)
    \\

    mgong081@connect.hkust-gz.edu.cn
}

\begin{document}

\maketitle

\begin{abstract}
    Accurate forecasting of spatiotemporal data remains challenging due to complex spatial dependencies and temporal dynamics. The inherent uncertainty and variability in such data often render deterministic models insufficient, prompting a shift towards probabilistic approaches, where diffusion-based generative models have emerged as effective solutions. In this paper, we present ProGen, a novel framework for probabilistic spatiotemporal time series forecasting that leverages Stochastic Differential Equations (SDEs) and diffusion-based generative modeling techniques in the continuous domain. By integrating a novel denoising score model, graph neural networks, and a tailored SDE, ProGen provides a robust solution that effectively captures spatiotemporal dependencies while managing uncertainty. Our extensive experiments on four benchmark traffic datasets demonstrate that ProGen outperforms state-of-the-art deterministic and probabilistic models. This work contributes a continuous, diffusion-based generative approach to spatiotemporal forecasting, paving the way for future research in probabilistic modeling and stochastic processes.
\end{abstract}

%

\section{Introduction}

Spatiotemporal data is always characterized by complex spatial dependencies and temporal dynamics, making accurate forecasting a challenging task. Many established time series forecasting models \cite{bai2020,lan2022,choi2023} often focus on deterministic predictions, which fail to capture the inherent uncertainty in the data.
Probabilistic models, which account for uncertainty in spatiotemporal data, are better suited to capture its complexities by generating a distribution of possible future outcomes, enabling more robust decision-making.
The rise of generative artificial intelligence has been significant in various fields \cite{meijer2024}.
Particularly, score-based generative models, also known as diffusion models,
which are a prominent category of probabilistic generative models originating from the field of image generation, have emerged as a promising approach.
They handle complex stochastic processes well \cite{yang2024}, making them ideal for spatiotemporal data due to their dynamic, interdependent capabilities.

In this work, we introduce ProGen: a novel framework for probabilistic spatial-temporal time series forecasting from a generative perspective using stochastic differential equations (SDEs). ProGen leverages the mathematical elegance of SDEs to model the evolution of spatiotemporal data as a continuous-time stochastic process.
Overall, it firstly perturbs the data into a Gaussian distribution through a forward diffusion process that involves training a model, and then iteratively denoises the data to sample predictions from a learnt distribution with the trained model.
ProGen redefines the model training, offering another solution for spatiotemporal forecasting with SDEs from a continuous-time generative modeling perspective.
It aims to address the current gap in probabilistic spatiotemporal time series forecasting. Unlike recent diffusion-based works in time series prediction \cite{rasul2021, yan2021a, bilos2023}, ProGen significantly improves the efficiency of generative models for sequence-to-sequence forecasting by taking non-autoregressive approach and considering spatial dependencies among different locations in the dataset. Also, distinct from \citet{wen2023},
ProGen operates in a continuous-time space, providing a more conceptually appropriate framework for time series data that aligns with the inherent continuity of real-world processes to effectively capture their evolving nature.
Key contributions include:

\begin{itemize}
    \item Conceptually, we propose a novel framework for probabilistic spatiotemporal graph time series forecasting from a continuous-time generative modeling perspective. We smoothly transform the data to provide predictions.
    \item Technically, we develop an innovative denoising network and a tailored SDE that account for spatiotemporal correlations within the data, enhancing the existing family of networks for spatiotemporal time series data. We have proved the sound theoretical foundation of the SDE and demonstrated the effectiveness of the entire network.
    \item Empirically, we demonstrate the superiority of ProGen over state-of-the-art deterministic and probabilistic models through extensive experiments on widely recognized real-world traffic datasets.
\end{itemize}

\section{Related Work}
\subsubsection{Diffusion Models.}
\citet{ho2020} and \citet{song2021} pioneered the development of diffusion models, focusing on generating data samples through forward diffusion and reverse denoising processes. The training image data \( x \) is viewed as drawn from a distribution \( p(x) \).
During the forward process, data is gradually transformed into noise, and the objective is to learn a distribution \( q_\theta(x) \) that approximates \( p(x) \). This enables the generation of new, unseen data samples during the reverse process \cite{luo2022}.
\citet{ho2020} detail the two processes as a series of discrete steps, whereas \citet{song2021} extend these processes to a continuous framework with SDEs.
Subsequent works have expanded these models to include conditional generation, allowing for the creation of targeted samples by incorporating specific attributes or guidance during the reverse process to steer outputs towards desired outcomes \cite{nichol2021, dhariwal2021, meijer2024}.

\subsubsection{Probabilistic Time Series Forecasting.}
With advancements in generative models, diffusion models have been adapted for time series forecasting. Models like TimeGrad \cite{rasul2021} and ScoreGrad \cite{yan2021a} utilize recurrent neural networks to encode history data and integrate it into the forward diffusion process.
However, they face challenges with slow generation speeds due to their autoregressive nature.
Also, the simplicity of the history encoders limits long-term forecasting accuracy \cite{shen2023}.
While most new diffusion models for time series utilize discrete approaches \cite{tashiro2021, wang2023c, chang2024, alcaraz2023a, wen2023},
ProGen employs a continuous method, adding noise directly and smoothly to data points and encodes history data more sophisticatedly—unlike the model by \citet{bilos2023}, which applies noise to continuous functions.
This approach improves the modeling of spatiotemporal correlations through advanced denoising networks and SDEs in the continuous domain.

\subsubsection{Spatiotemporal Forecasting.}
Well-performing spatiotemporal forecasting methods include AGCRN \cite{bai2020} and DSTAGNN \cite{lan2022}, crafting dynamic graphs among nodes; and STG-NRDE \cite{choi2023}, leveraging neural rough differential equations. However, these deterministic models lack probabilistic consideration. While discrete diffusion models like DiffSTG \cite{wen2023} have ventured into probabilistic forecasting, they maintain a discrete approach and do not capture the continuous nature of time series.


\section{Preliminaries}

\subsubsection{Problem Setup.} We address probabilistic spatiotemporal time series forecasting, aiming to predict future values of a spatiotemporal series using historical data. Formally, let \(\mathcal{D} = \{\mathbf{X_t}\}_{t=1}^T\) be a spatiotemporal dataset, where \(\mathbf{X_t} \in \mathbb{R}^{N \times D}\) denotes observations at time \(t\), with \(N\) spatial locations and \(D\) features. Spatial dependencies among the \(N\) locations are represented by a graph \(\mathcal{G} = (\mathcal{V}, \mathcal{E}, A)\), where \(\mathcal{V}\) is the set of nodes (locations), \(\mathcal{E}\) the set of edges (connections), and \(A \in \mathbb{R}^{N \times N}\) the adjacency matrix encoding the graph's connectivity.
The probabilistic prediction task involves estimating the distribution \(q_X (\mathbf{X_{T+1:T+H}} \mid \mathbf{X_{T-L+1:T}}, \mathcal{G}, \mathcal{C})\), where \(\mathbf{L}\) is the historical window length, \(\mathbf{H}\) the forecasting horizon, and \(\mathcal{C}\) other covariates, using a sequence-to-sequence approach. For clarity, \(\mathbf{X_{F}}\) denotes future ground truth \(\mathbf{X_{T+1:T+H}}\), \(\mathbf{X_{H}}\) denotes historical observations \(\mathbf{X_{T-L+1:T}}\), and \(\mathbf{P_{H}}\) denotes temporal markers such as week and day positions within the historical data.

\subsubsection{Stochastic Differential Equations.} SDEs are a powerful mathematical framework for modeling continuous-time stochastic processes. An SDE is defined as follows:
\begin{equation}
    d\mathbf{X} = f(\mathbf{X}, t)dt + g(\mathbf{X}, t)dW,
    \label{eq:sde}
\end{equation}
where \(\mathbf{X} \in \mathbb{R}^d\) is the state of the system at diffusion timestep \(t\), with \(t\) indexed discretely by \(\left\{0, \frac{1}{K}, \frac{2}{K}, \ldots, \frac{K-1}{K}, 1\right\}\) to represent the subdivision of the continuous interval \([0, 1]\) into \(K\) steps. Here, \(d\) represents the dimensionality of the state space. \(f: \mathbb{R}^d \times \mathbb{R} \rightarrow \mathbb{R}^d\) is the drift function, which describes the deterministic part of the system's evolution. \(g: \mathbb{R}^d \times \mathbb{R} \rightarrow \mathbb{R}^{d \times m}\) is the diffusion function, which models the stochastic influence, with \(m\) representing the number of independent Brownian motion components. \(W\) denotes a standard \(m\)-dimensional Brownian motion.


\subsubsection{Reverse Stochastic Differential Equations.} To recover original data from noisy states, we use the reverse SDE, describing the dynamics of the reverse process as per \citet{anderson1982}. It is defined as:
\begin{equation}
    \begin{aligned}
        d\mathbf{X} = & \left[f(\mathbf{X}, t) - g^2\mathbf{(X}, t)\nabla_\mathbf{X} \log p_t(\mathbf{X})\right]dt  + g(\mathbf{X}, t)d\bar{W},
    \end{aligned}
    \label{eq:rsde}
\end{equation}
where \(\nabla_\mathbf{X} \log p_t(\mathbf{X})\) is the score function (gradient of the log probability density) at diffusion timestep \(t\), and \(\bar{W}\) is the reverse Wiener process. This reverse SDE iteratively denoises the data, retrieving the original distribution.

\subsubsection{Denoising Score Matching (DSM).}
DSM is a score-based generative modeling approach that learns by minimizing the discrepancy between the score functions of the model and data distributions across diffusion time steps. The DSM objective is:

\begin{equation}
    \begin{split}
        \mathcal{L}(\theta) = \mathbb{E}_{t \sim \text{Uniform}(0, 1)} \mathbb{E}_{X \sim p_{\text{data}}}[\|\nabla_X \log q_{\theta}(\mathbf{X^t} | t) \\
        - \nabla_X \log p_{\text{data}}(\mathbf{X^t} | t)\|^2],
    \end{split}
\end{equation}
where $q_{\theta}(\mathbf{X^t} | t)$ and $p_{\text{data}}(\mathbf{X^t} | t)$ are the model and data distributions at $t$, and $\theta$ includes all model parameters.

\section{Methodology}

\begin{figure}[ht!]
    \centering
    \includegraphics[width=0.49\textwidth]{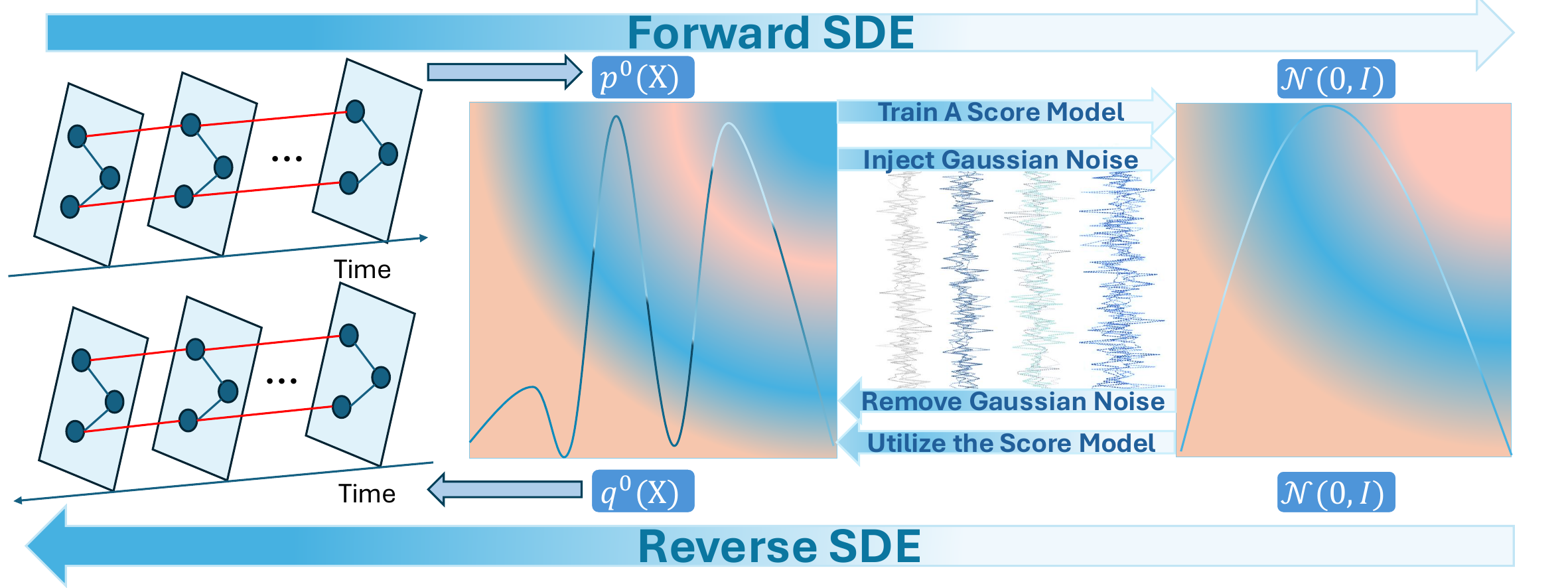}
    \caption{Overview of The Two Processes of ProGen for Probabilistic Spatiotemporal Time Series Forecasting.}
    \label{fig:framework}
\end{figure}

As shown in Fig~\ref{fig:framework}, ProGen operationalizes probabilistic forecasting through two principal processes:
First, the forward diffusion process, where future ground truths $\mathbf{X_F}$ from the training dataset are transformed into a Gaussian state $\mathcal{N}(0, I)$, concurrently with the training of a score model that learns from this transformation.
Second, a reverse prediction process that iteratively denoises $N$ samples from this Gaussian state across \(K\) discretized timesteps, defined as \(t = \frac{k}{K}\) for \(k \in \{K, K-1, \ldots, 0\}\). This reverse process generates the forecasting distribution, directed by the trained score model's estimates, \( \nabla \log p(\mathbf{X}^t | \mathbf{X_H}) \), which navigates the recovery towards the highest probability future state \( \mathbf{X}_\mathbf{F}^0 \), contingent on \( \mathbf{X_H} \), at each diffusion timestep \( t \).

\subsection{Forward Diffusion Process}
In the forward diffusion process, \(\mathbf{X_{F}}\) from the training dataset is gradually perturbed by Gaussian noise, aiming to transform the data towards an isotropic Gaussian distribution, simultaneously facilitating the training of a score model. Governed by the SDE described in Equation~\ref{eq:sde}, this process perturbs each data point at diffusion timesteps \( t \), which are drawn uniformly from the interval [0, 1], representing continuous time steps during the diffusion. The perturbation at each timestep is characterized by:
\begin{equation}
    \mathbf{\tilde{X}^t_{F}} = \mu(\mathbf{X_{F}}, t; \beta_0, \beta_1) + \sigma(\mathbf{X_{F}}, t; \beta_0, \beta_1) \times Z,
\end{equation}
where \( Z \sim \mathcal{N}(0, I) \) denotes standard Gaussian noise. The functions \(\mu\) and \(\sigma\), parameterized by \(\beta_0\) and \(\beta_1\), compute the mean and standard deviation of the perturbed data at each timestep \( t \), transforming the original data distribution \( p_{\mathbf{X}}^0(\mathbf{X_{F}}) \) into a noise-dominated distribution \( q_\mathbf{X}^t(\mathbf{X_{F}}) \) across the continuous diffusion timeframe.


\subsection{Training Denoising Score Model}

\begin{figure*}[ht]
    \centering
    \includegraphics[width=0.75\textwidth]{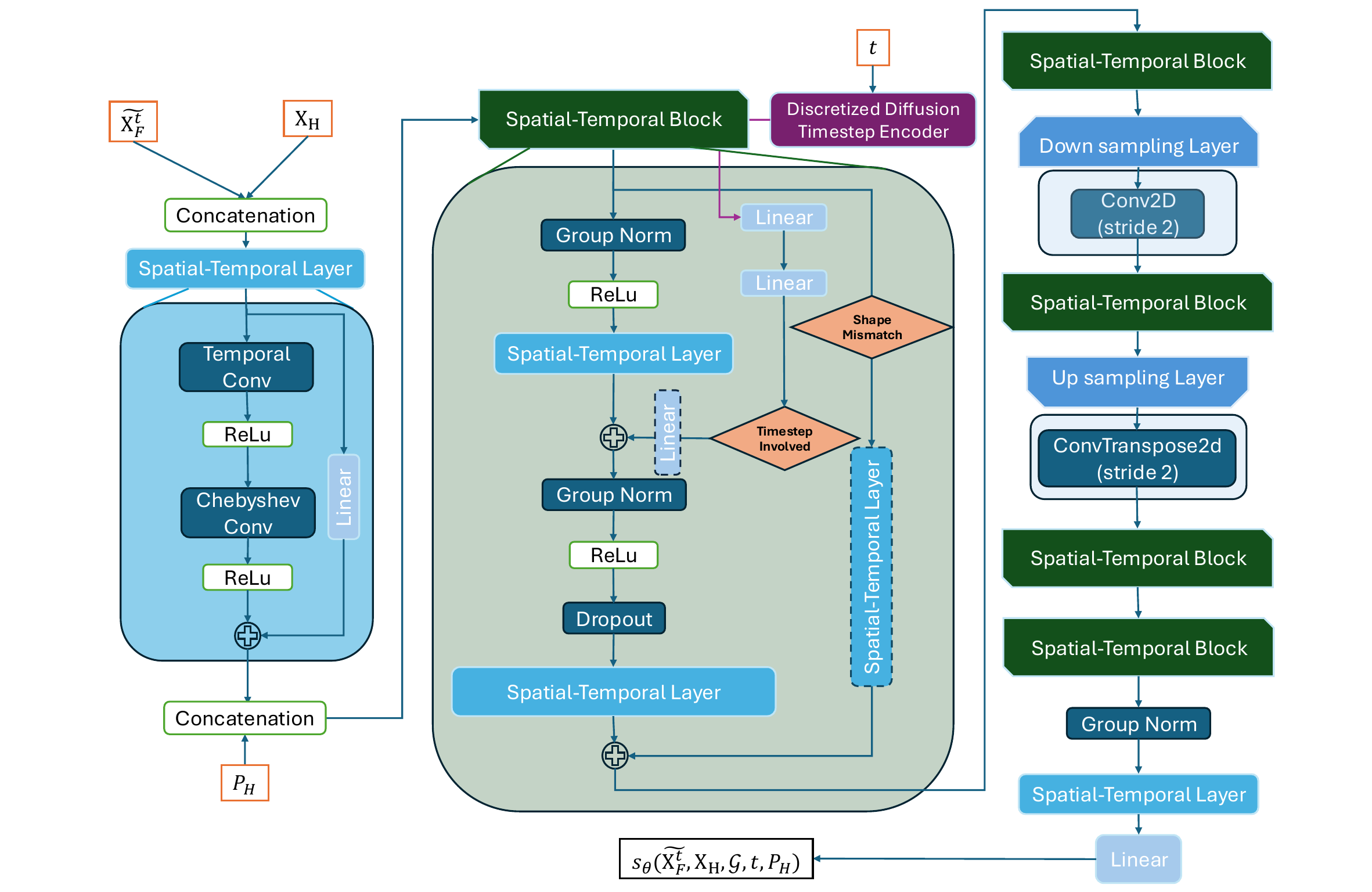}
    \caption{Architecture of the Denoising Score Matching Model in ProGen.}
    \label{fig:model}
\end{figure*}

The score model \(s_\theta(\mathbf{\tilde{X}^t_{F}}, \mathbf{X_{H}}, \mathcal{G}, t, \mathbf{P_{H}})\) in ProGen is pivotal for estimating the scores (gradients of the log probability density function), which are critical for the reverse denoising process. This model is trained through a loss function designed to minimize the difference between estimated and true gradients, as described by:

{\footnotesize
\begin{equation}
    \begin{aligned}
        \mathcal{L}(\theta) & = \mathbb{E}_{t} \left\{ \lambda(t) \mathbb{E}_{\mathbf{X}_\mathbf{F}^0, \mathbf{X_H}} \left[ \mathbb{E}_{\tilde{\mathbf{X}}_\mathbf{F}^t | \mathbf{X}_\mathbf{F}^0, \mathbf{X_H}} \right. \right.                                                                         \\
                            & \quad \left. \left. \left\| s_\theta(\tilde{\mathbf{X}}_\mathbf{F}^t, \mathbf{X_H}, \mathcal{G}, t, \mathbf{P_H}) - \nabla_{\tilde{\mathbf{X}}_\mathbf{F}^t} \log p(\tilde{\mathbf{X}}_\mathbf{F}^t | \mathbf{X}_\mathbf{F}^0, \mathbf{X_H}) \right\|_2^2 \right] \right\}
    \end{aligned}
    \label{eq:loss}
\end{equation}
}

Here, \(\lambda : [0, K] \rightarrow \mathbb{R}_{>0}\) represents a positive weighting function.
\(\tilde{\mathbf{X}}_\mathbf{F}\) are sampled from the conditional distribution \(p^{0t}
\). With sufficient data and model capacity, score matching ensures that the optimal solution to the objective function, denoted by \(s_\theta
\), approximates \(\nabla_{\tilde{{X}}_\mathbf{F}} \log p^t
\) for almost all \(\tilde{\mathbf{X}}_\mathbf{F}\) across \(t \in [0, 1]\). Similar to \citet{song2021}, we  choose \(\lambda\) proportional to \(
\mathbb{E}[||\nabla_{\tilde{{X}}_\mathbf{F}^t} \log p^{0t}
    ||_2^2]
\).

As shown in Fig~\ref{fig:model}, $\tilde{\mathbf{X}}_\mathbf{F}^t$ and $\mathbf{X}_\mathbf{H}$ are concatenated, incorporating historical context to enhance score estimation for the reverse diffusion process. This methodology diverges from traditional denoising score models in image generation \cite{song2021, ho2020, nichol2021, dhariwal2021} and contrasts with recent discrete diffusion applications in spatiotemporal forecasting \cite{wen2023}.
The input traverses spatial-temporal layers arranged in blocks that include temporal and Chebyshev convolutions, optimizing for both spatial and temporal dependencies. Only the first block's initial layer conditions on \(t\), as per the encoding method in \cite{ho2020}, enhancing robustness and training efficiency through a residual setup.
A dual-scale strategy involving downsampling and subsequent upsampling allows the model to accurately detect broad patterns and detailed aspects of high-dimensional spatiotemporal data. This refined architecture facilitates precise gradient estimation, essential for effectively training the model, as detailed in Algorithm~\ref{alg:train}.

\begin{algorithm}[tb]
    \caption{Training Algorithm for ProGen}
    \label{alg:train}
    \textbf{Input}: Future ground truths in the training set $\mathbf{X_{F}}$, historical observations $\mathbf{X_{H}}$, graph structure $\mathcal{G}$, positions of week and day information in history data $\mathbf{P_{H}}$ \\
    \textbf{Parameter}: Number of epochs $E$
    \\
    \textbf{Output}: Trained score model $s_\mathcal{\theta^*}$
    \begin{algorithmic}[1]
        \FOR{epoch $= 1$ to $E$}
        \FOR{each sample in the training set}
        \STATE Initialize discretized timestep $t \sim \mathcal{U}(\mathbf{0}, \mathbf{I})$, random noise $\mathbf{Z} \sim \mathcal{N}(\mathbf{0}, \mathbf{I})$
        \STATE Compute mean $\mathbf{m^t}$ and standard deviation $\mathbf{\sigma^t}$ of marginal probability \(p^{0t}(\tilde{\mathbf{X}}_\mathbf{F}^t | \mathbf{X}_\mathbf{F})\)
        \STATE Generate perturbed samples $\tilde{\mathbf{X}}_{F}^t \leftarrow \mathbf{m^t} + \mathbf{\sigma^t} \cdot \mathbf{Z}$
        \STATE Compute loss $\mathcal{L}(\theta)$ with $\mathbf{X_{F}}$, $\mathbf{t}$, $\mathcal{G}$ and  $\mathbf{P_{H}}$ as in Equation~\ref{eq:loss}
        \STATE Take the gradient of the loss $\nabla_{\theta} \mathcal{L}(\theta)$ and update model parameters

        \ENDFOR
        \ENDFOR
        \STATE \textbf{return} $s_\mathcal{\theta^*}$
    \end{algorithmic}
\end{algorithm}

\subsection{Adaptive Reverse Prediction Process}
The forward diffusion process (training phase) exposes the model to increasingly perturbed data points towards the state \(\mathcal{N}(0, {I})\).
The reverse prediction process in ProGen iteratively restores the original data distribution from \(N\) sample points initialized from \(\mathcal{N}(0, I)\) at the final \(K\)-th diffusion timestep.
Governed by the reverse SDE in Equation \ref{eq:rsde}, this process spans \(K\) discretized timesteps from \(t=1\) to \(t=0\) (\(t = \frac{k}{K}\), where \(k\) decrements from \(K\) to 0), refining each sample based on estimated gradients.
To address the intricacies of spatiotemporal data, we utilize a spatiotemporal SDE (ST SDE), as outlined in Equation \ref{eq:spatiotemporal_sde}, which extends the sub-VP SDE framework \cite{song2021}. This ST SDE incorporates spatial relationships via an adjacency matrix \(A\), enhancing the drift coefficient to better capture spatial interactions:

{\footnotesize
\begin{equation}
    \begin{aligned}
        d\mathbf{X} & = -\frac{1}{2} \beta(t)\left( \mathbf{X} - \alpha A \mathbf{X} \right)dt  + \sqrt{\beta(t)(1 - e^{-2 \int_0^t \beta(s) \, ds})} \, dw,
    \end{aligned}
    \label{eq:spatiotemporal_sde}
\end{equation}
}

Specifically, at each reverse diffusion step \( t \), the approximation of the original data distribution for each sample is updated as follows:

{\footnotesize
\begin{equation}
    \begin{aligned}
        \tilde{\mathbf{X}}_{\mathbf{F}}^{t - \frac{1}{K}} & = \tilde{\mathbf{X}}_{\mathbf{F}}^{t} - \left[ -\frac{1}{2} \beta(t) \left( \tilde{\mathbf{X}}_{\mathbf{F}}^{t} - \alpha A \tilde{\mathbf{X}}_{\mathbf{F}}^{t} \right) \, dt \right. \\
                                                          & \quad \left. - \left(\sqrt{\beta(t)(1 - e^{-2 \int_0^{t} \beta(s) \, ds})} \right)^2 \times s_\theta \right]                                                                         \\
                                                          & \quad + \sqrt{\beta(t)(1 - e^{-2 \int_0^{t} \beta(s) \, ds})} \cdot dW,
    \end{aligned}
    \label{eq:reverse_stdse}
\end{equation}
}

The proposed ST SDE retains the core structural properties of the sub-VP SDE and VP SDE described in \cite{song2021} with neighborhood influences in the drift coefficient, enhancing spatial interaction modeling.
Notably, we observed that ST SDE converges more quickly than the sub-VP SDE.
This accelerated convergence, while enhancing responsiveness to spatial dependencies, tends to correct perturbations more aggressively. Such behavior increases the risk of premature convergence, potentially bypassing nuanced equilibrium states essential for achieving the optimally denoised solution.
To mitigate this, we introduce an adaptive mechanism that adjusts between ST SDE and sub-VP SDE based on performance metrics, optimizing the denoising across diffusion steps. We prove that the ST SDE maintains convergence properties akin to its predecessors, with detailed proofs provided in \textcolor{black}{Appendix A}. Algorithms about the entire adaptive reverse workflow and specific reverse sampling processes are detailed in the \textcolor{black}{Appendix E}.

\section{Experiments}

\subsubsection{Dataset Overview.}
We conducted experiments using four traffic benchmark datasets provided by the California Performance Measurement System (PEMS) \cite{chen2001}. These datasets—PEMS03, PEMS04, PEMS07, and PEMS08—were released by \cite{song2020a}. Each dataset includes traffic flow data collected from various sensors installed along highways. The spatial adjacency graph for each dataset is constructed based on the corresponding road network. Statistics of these datasets are in \textcolor{black}{Appendix C}.

\begin{table*}[htb]
    \centering
    \renewcommand{\arraystretch}{1.1} 
    \begin{adjustbox}{max width=\textwidth}
        \begin{tabular}{c|c|c|c|c|c|c|c|c|c|c||c}
            \hline
            \textbf{Datasets}       & \textbf{Metric} & \textbf{AGCRN} & \textbf{STSGCN} & \textbf{DSTAGNN} & \textbf{STGNCDE} & \textbf{STGNRDE} & \textbf{ARIMA} & \textbf{FCLSTM} & \textbf{MTGNN} & \textbf{ASTGCN(r)} & \textbf{ProGen} \\ \hline
            \multirow{2}{*}{PEMS03} & MAE             & 15.98          & 17.48           & 15.57            & 15.57            & 15.50            & 35.41          & 21.33           & 16.46          & 17.34              & \textbf{15.07}  \\ \cline{2-12}
                                    & RMSE            & 28.25          & 29.21           & 27.21            & 27.09            & 27.06            & 47.59          & 35.11           & 28.56          & 29.56              & \textbf{25.09}  \\ \cline{2-12}
            \hline

            \multirow{2}{*}{PEMS08} & MAE             & 15.95          & 17.13           & 15.67            & 15.45            & 15.32            & 31.09          & 23.09           & 15.71          & 18.25              & \textbf{14.99}  \\ \cline{2-12}
                                    & RMSE            & 25.22          & 26.80           & 24.77            & 24.81            & 24.72            & 44.32          & 35.17           & 24.62          & 28.06              & \textbf{24.00}  \\ \cline{2-12}
            \hline
        \end{tabular}
    \end{adjustbox}
    \caption{Deterministic performance on full PEMS03 and PEMS08 datasets.}
    \label{table:det-metrics}
\end{table*}

\begin{table*}[htb]
    \centering
    \renewcommand{\arraystretch}{1.1} 
    \begin{adjustbox}{max width=\textwidth}
        \begin{tabular}{c|c|c|c|c|c|c|c||c}
            \hline
            \textbf{Datasets}       & \textbf{Metric} & \textbf{AGCRN*} & \textbf{DSTAGNN*} & \textbf{STGNRDE*} & \textbf{DeepAR} & \textbf{DiffSTG} & \textbf{CSDI} & \textbf{ProGen} \\ \hline
            \multirow{4}{*}{PEMS03} & MAE             & 16.79           & 17.03             & 25.40             & 23.76           & 53.52            & 24.52         & \textbf{14.14}  \\ \cline{2-9}
                                    & RMSE            & 29.26           & 31.05             & 37.93             & 37.82           & 69.62            & 38.93         & \textbf{22.71}  \\ \cline{2-9}
                                    & CRPS            & \diagbox{}{}    & \diagbox{}{}      & \diagbox{}{}      & 0.11            & 0.24             & 0.11          & \textbf{0.07}   \\ \cline{2-9}
                                    & MIS             & \diagbox{}{}    & \diagbox{}{}      & \diagbox{}{}      & 205.78          & 466.97           & 245.86        & \textbf{106.50} \\ \hline
            \multirow{4}{*}{PEMS04} & MAE             & 21.60           & 21.25             & 30.67             & 29.59           & 32.55            & 27.83         & \textbf{18.87}  \\ \cline{2-9}
                                    & RMSE            & 34.32           & 32.82             & 44.29             & 46.01           & 46.89            & 42.71         & \textbf{30.01}  \\ \cline{2-9}
                                    & CRPS            & \diagbox{}{}    & \diagbox{}{}      & \diagbox{}{}      & 0.10            & 0.11             & 0.10          & \textbf{0.07}   \\ \cline{2-9}
                                    & MIS             & \diagbox{}{}    & \diagbox{}{}      & \diagbox{}{}      & 237.75          & 209.85           & 223.77        & \textbf{138.88} \\ \hline
            \multirow{4}{*}{PEMS07} & MAE             & \textbf{21.50}  & 24.06             & 33.48             & 28.77           & 37.80            & 30.57         & 21.91           \\ \cline{2-9}
                                    & RMSE            & 36.08           & 39.02             & 46.49             & 44.80           & 50.31            & 45.92         & \textbf{35.11}  \\ \cline{2-9}
                                    & CRPS            & \diagbox{}{}    & \diagbox{}{}      & \diagbox{}{}      & 0.07            & 0.09             & 0.07          & \textbf{0.05}   \\ \cline{2-9}
                                    & MIS             & \diagbox{}{}    & \diagbox{}{}      & \diagbox{}{}      & 242.61          & 293.45           & 262.15        & \textbf{186.78} \\ \hline
            \multirow{4}{*}{PEMS08} & MAE             & 16.90           & 15.78             & 25.10             & 23.15           & 44.47            & 19.00         & \textbf{15.46}  \\ \cline{2-9}
                                    & RMSE            & 26.47           & \textbf{24.36}    & 36.32             & 35.92           & 60.72            & 28.99         & 24.71           \\ \cline{2-9}
                                    & CRPS            & \diagbox{}{}    & \diagbox{}{}      & \diagbox{}{}      & 0.08            & 0.15             & 0.07          & \textbf{0.05}   \\ \cline{2-9}
                                    & MIS             & \diagbox{}{}    & \diagbox{}{}      & \diagbox{}{}      & 191.20          & 284.64           & 146.06        & \textbf{120.53} \\ \hline
        \end{tabular}
    \end{adjustbox}
    \caption{Overall performance on all datasets. *CRPS and MIS metrics are not applicable to deterministic models.}
    \label{table:sup_metrics}
\end{table*}

\begin{table}[ht]
    \centering
    \begin{tabular}{l|c|c|c}
        \hline
        \textbf{Method} & \textbf{MAE}   & \textbf{RMSE}  & \textbf{CRPS} \\
        \hline
        Latent ODE      & 26.05          & 39.50          & 0.11          \\        \hline
        DeepAR          & 21.56          & 33.37          & 0.07          \\        \hline
        CSDI            & 32.11          & 47.40          & 0.11          \\        \hline
        TimeGrad        & 24.46          & 38.06          & 0.09          \\        \hline
        MC Dropout      & 19.01          & 29.35          & 0.07          \\        \hline
        DiffSTG         & 18.60          & 28.20          & 0.06          \\        \hline
        \hline
        \textbf{ProGen} & \textbf{14.99} & \textbf{24.00} & \textbf{0.05} \\
        \hline
    \end{tabular}
    \caption{Probabilistic performance on full PEMS08 dataset.}
    \label{table:prob-metrics}
\end{table}

\subsubsection{Baselines.}

To validate ProGen's effectiveness in both deterministic and probabilistic forecasting, we compare it against a spectrum of traditional and advanced models. Deterministic baselines include \textbf{ARIMA}, \textbf{FCLSTM} \cite{sutskever2014}, \textbf{MTGNN} \cite{wu2020}, \textbf{ASTGCN} \cite{zhu2020}, \textbf{AGCRN} \cite{bai2020}, \textbf{STSGCN} \cite{song2020a}, \textbf{DSTAGNN} \cite{lan2022}, \textbf{STGNCDE} \cite{choi2022}, and \textbf{STGNRDE} \cite{choi2023}. These models range from classic ARIMA to complex graph-based networks that integrate dynamic spatial-temporal patterns and differential equations. Probabilistic baselines include \textbf{Latent ODE} \cite{rubanova2019}, \textbf{DeepAR} \cite{salinas2020}, \textbf{CSDI} \cite{tashiro2021}, \textbf{TimeGrad} \cite{rasul2021}, \textbf{MC Dropout} \cite{wu2021a}, and \textbf{DiffSTG} \cite{wen2023}, employing various advanced neural architectures and differential equations to manage uncertainties in time series data. Detailed descriptions of baselines are in \textcolor{black}{Appendix F1}.


\begin{figure*}[htb]
    \centering
    \begin{subfigure}{0.4\textwidth}
        \centering
        \includegraphics[width=\textwidth]{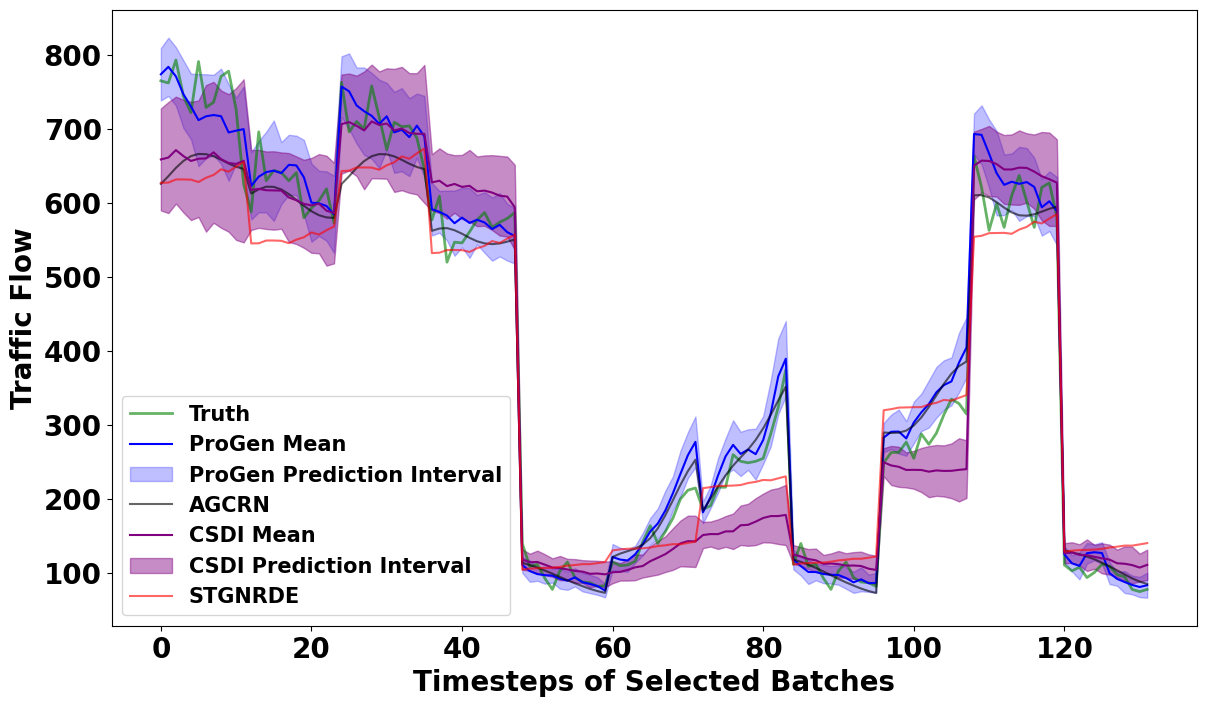}
        \caption{Node 223 in PEMS03.}
        \label{fig:node223_pems03}
    \end{subfigure}%
    \hspace{0.02\textwidth} 
    \begin{subfigure}{0.4\textwidth}
        \centering
        \includegraphics[width=\textwidth]{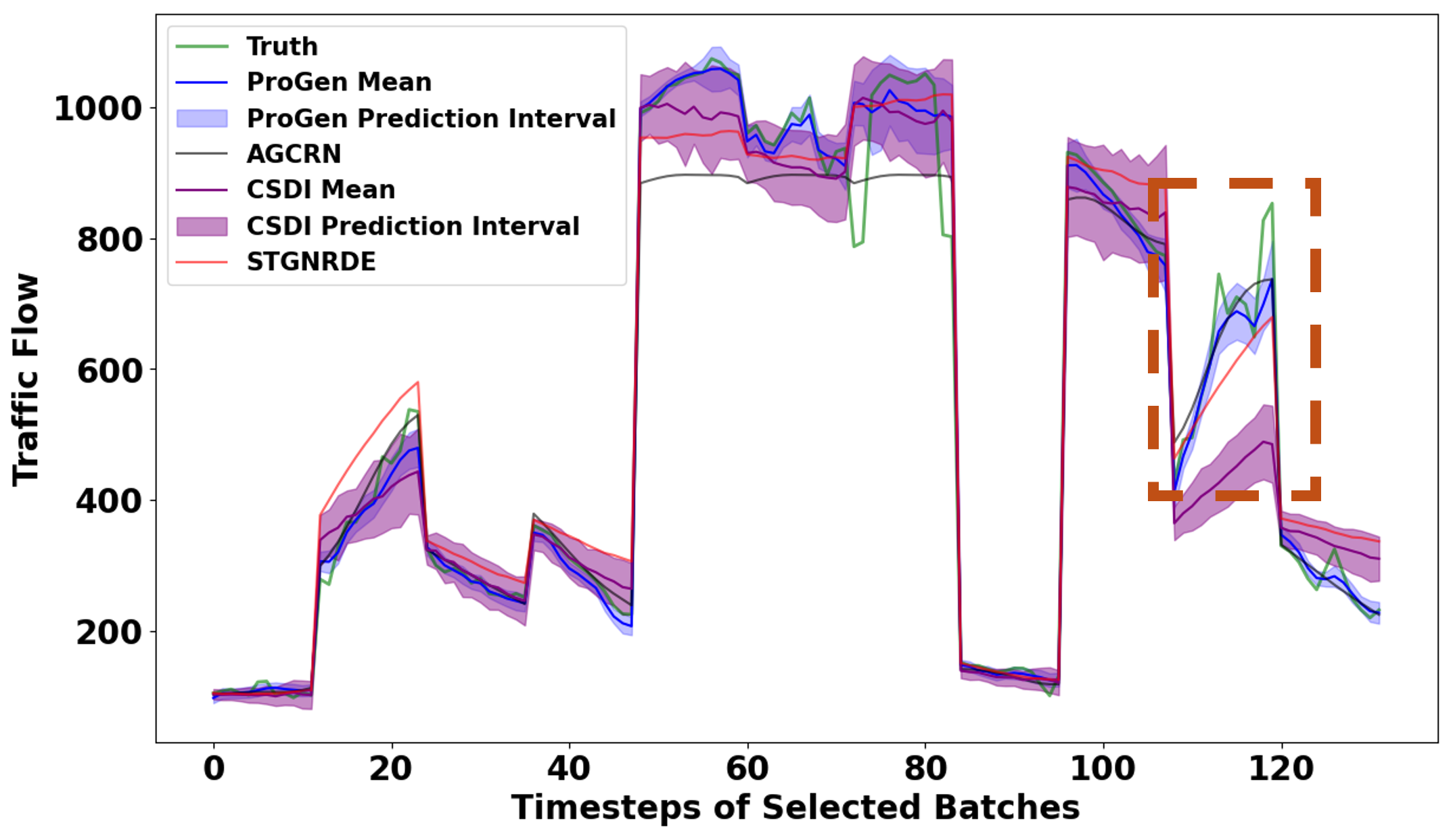}
        \caption{Node 537 in PEMS07.}
        \label{fig:node537_pems07}
    \end{subfigure}
    \caption{Visualization of traffic forecasting predictions among strong baselines.}
    \label{fig:pems03_04_07_08_nodes_vis}
\end{figure*}

\subsubsection{Evaluation Metrics.}
For deterministic forecasting, we measure accuracy using Mean Absolute Error (MAE) and Root Mean Squared Error (RMSE). For probabilistic forecasting, quality is assessed via Continuous Ranked Probability Score (CRPS) and Mean Interval Score (MIS). Lower scores in all metrics signify improved performance, denoting smaller deviations from actual values and more reliable forecasts. Details for each metric are in \textcolor{black}{Appendix D}.

\subsubsection{Hyperparameters.}
We divided the data into training, validation, and test sets in a 6:2:2 ratio, selecting the best model based on the lowest validation loss. ProGen's hyperparameters include: spatial structure influence on the drift term \(\alpha\) ranging from -0.3 to 0.9 (Equation~\ref{eq:spatiotemporal_sde}), spatial-temporal layer output dimension of \(\{32, 64, 128\}\), hidden dimensions ranging from 16 to 256, position embedding dimensions of day and week \(\{8, 16, 32, 64, 128\}\), number of residual blocks \(\{1, 2, 3, 4, 5\}\), channel multipliers for downsampling and upsampling \(\{[1, 2], [1, 2, 2], [1, 2, 3], [1, 2, 3, 4]\}\), learning rate from \(1\times10^{-4}\) to \(1\times10^{-2}\), and a batch size of 128. The Adam optimizer is used, and the model undergoes training for 300 epochs.
For baseline models, we use official dataset results when available. If not, we adhere to hyperparameters from the original publications, refining them on the validation set to ascertain optimal settings for test set evaluations. Detailed experiments setups are in \textcolor{black}{Appendix F2}.

\begin{figure*}[ht]
    \centering
    \includegraphics[width=0.8\textwidth]{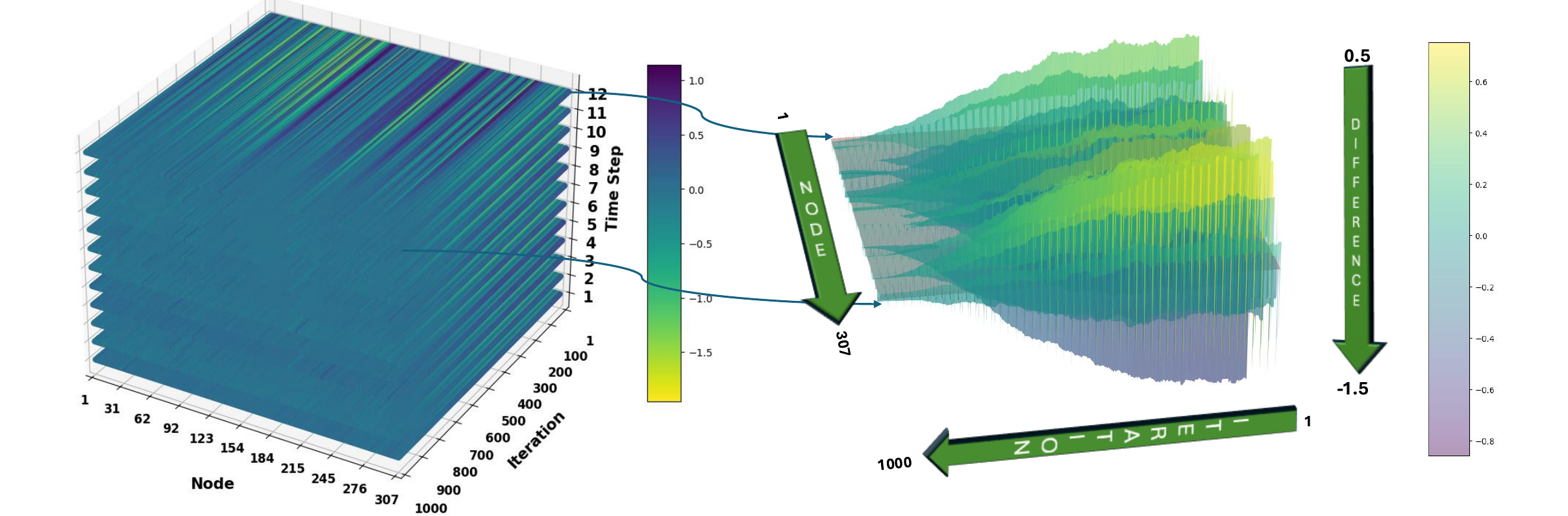}
    \caption{Visualization of the difference between the mean predictions and the actual values for the average test data in PEMS04.}
    \label{fig:pems04_3d_waves_plot}
\end{figure*}

\section{Results}
\begin{figure}[ht]
    \centering
    \includegraphics[width=0.4\textwidth]{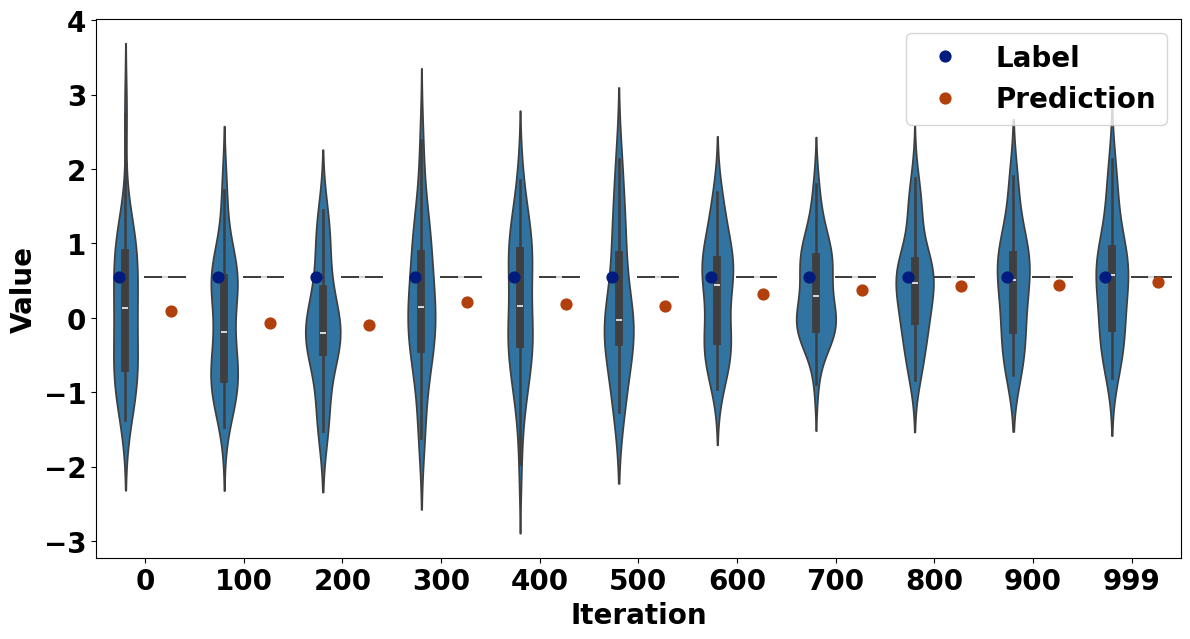}
    \caption{Distribution and mean values of predictions versus actual truths across initial batch iterations in PEMS04.}
    \label{fig:distribution_change_iteration}
\end{figure}

\begin{figure}[htbp]
    \centering
    \includegraphics[width=0.35\textwidth]{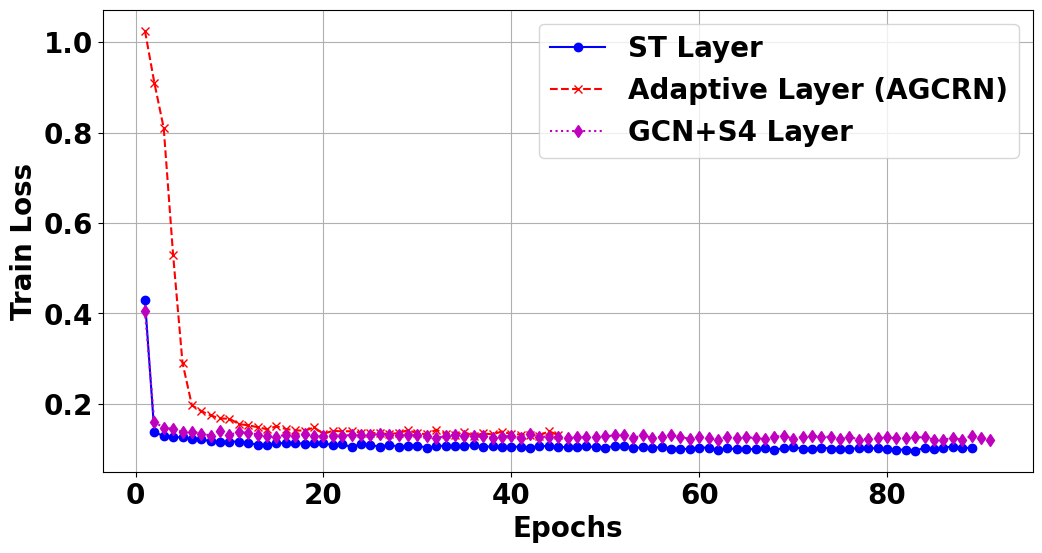}
    \caption{Training loss curves for different spatiotemporal layers in PEMS08 over 100 epochs.}
    \label{fig:train_loss}
\end{figure}

We conducted full test runs on the PEMS03 and PEMS08 datasets—PEMS03 having the most samples and PEMS08 being the most commonly used. For comprehensive comparisons, we performed random runs on all datasets, selecting 11 batches of data from the test set and running the models 10 times each. This approach strikes a balance between the computational intensity and time requirements of generative modeling tasks, and the need for a fair comparison. We assessed the statistical significance of these random run results using the Wilcoxon signed-rank and paired t-tests to determine if ProGen outperforms or underperforms the baselines beyond random variation. By conducting these significance tests, we ensure that the random test results are both reliable and valid, demonstrating its performance authentically.

\begin{figure}[htb]
    \centering
    \begin{subfigure}[b]{0.233\textwidth}
        \centering
        \includegraphics[width=\textwidth]{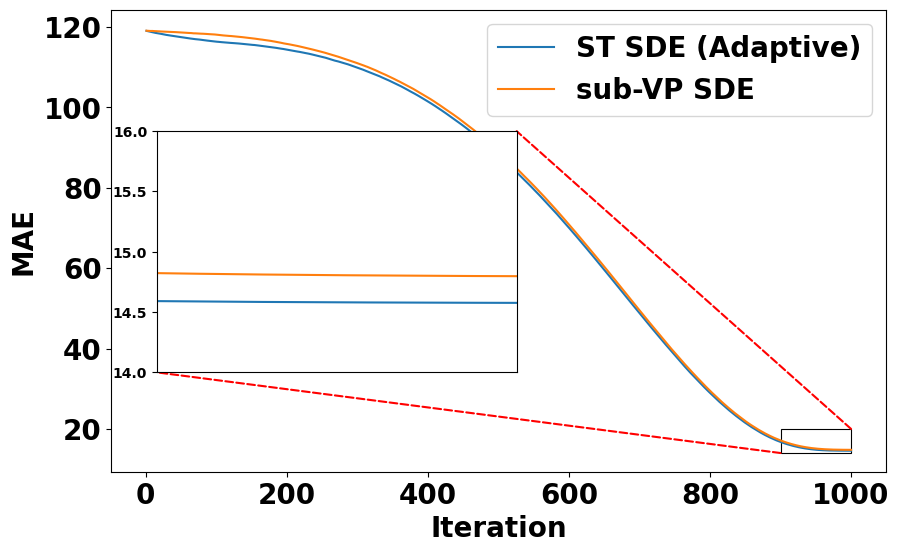}
        \caption{MAE comparison}
        \label{fig:mae_stsde_subvpsde}
    \end{subfigure}
    \hfill
    \begin{subfigure}[b]{0.233\textwidth}
        \centering
        \includegraphics[width=\textwidth]{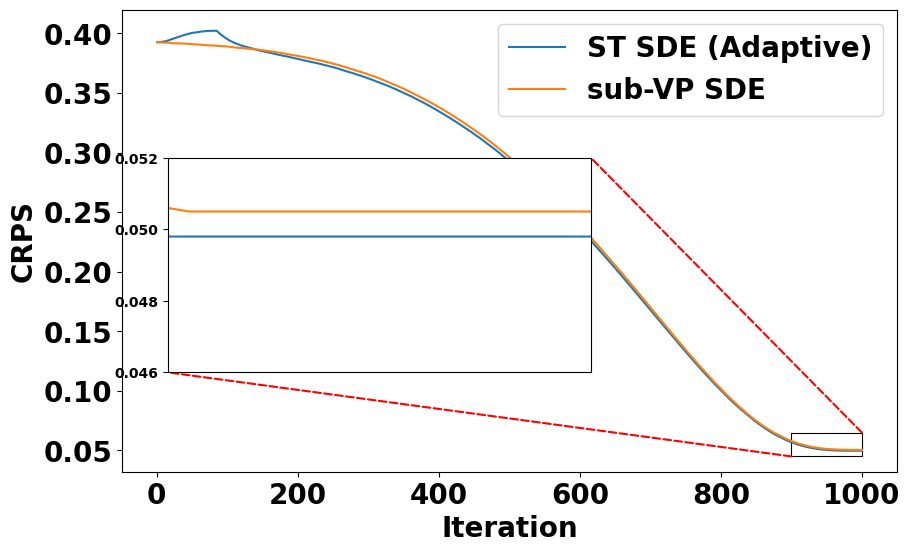}
        \caption{CRPS comparison }
        \label{fig:crps_stsde_subvpsde}
    \end{subfigure}
    \caption{Performance comparison between the tailored spatial-temporal SDE and the subVP-SDE in PEMS08.}
\end{figure}



\begin{figure}[htb]
    \centering
    \begin{subfigure}[b]{0.45\textwidth}
        \centering
        \includegraphics[width=\textwidth]{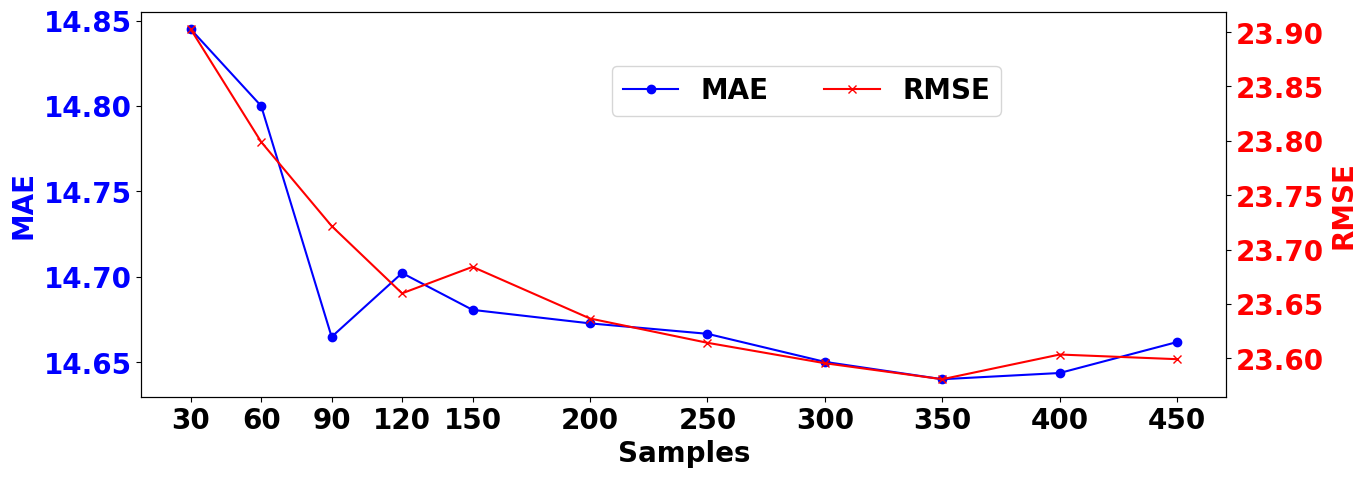}
        \caption{MAE and RMSE}
        \label{fig:mae_rmse_samples}
    \end{subfigure}
    \hfill
    \begin{subfigure}[b]{0.45\textwidth}
        \centering
        \includegraphics[width=\textwidth]{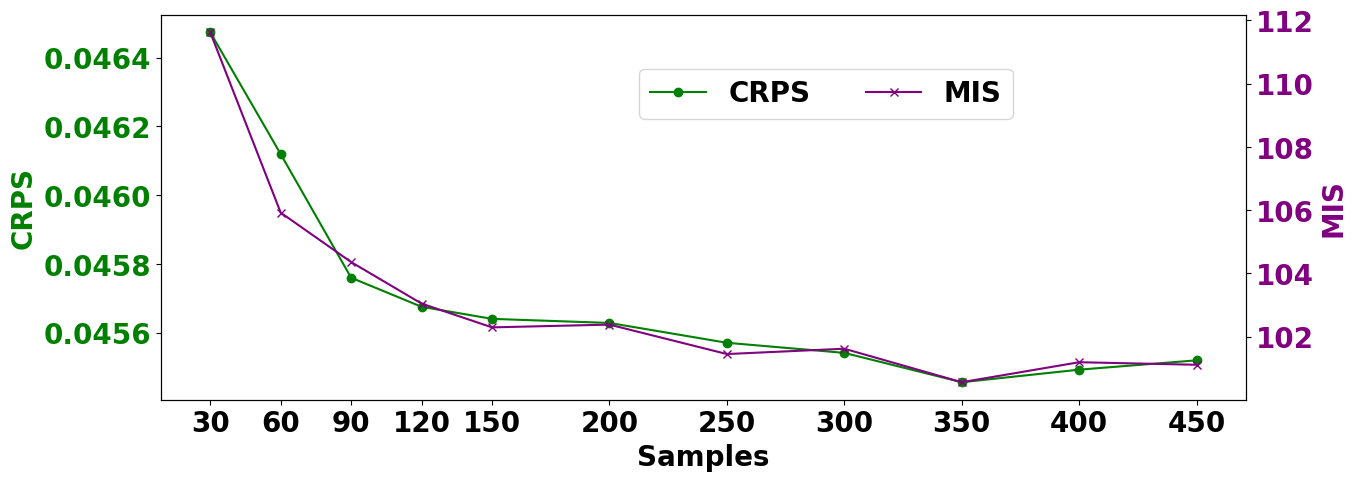}
        \caption{CRPS and MIS}
        \label{fig:crps_mis_samples}
    \end{subfigure}
    \caption{Metrics across varying sample sizes in PEMS08.}
    \label{fig:sample_size}
\end{figure}

\begin{figure}[htb]
    \centering
    \begin{subfigure}[b]{0.45\textwidth}
        \centering
        \includegraphics[width=\textwidth]{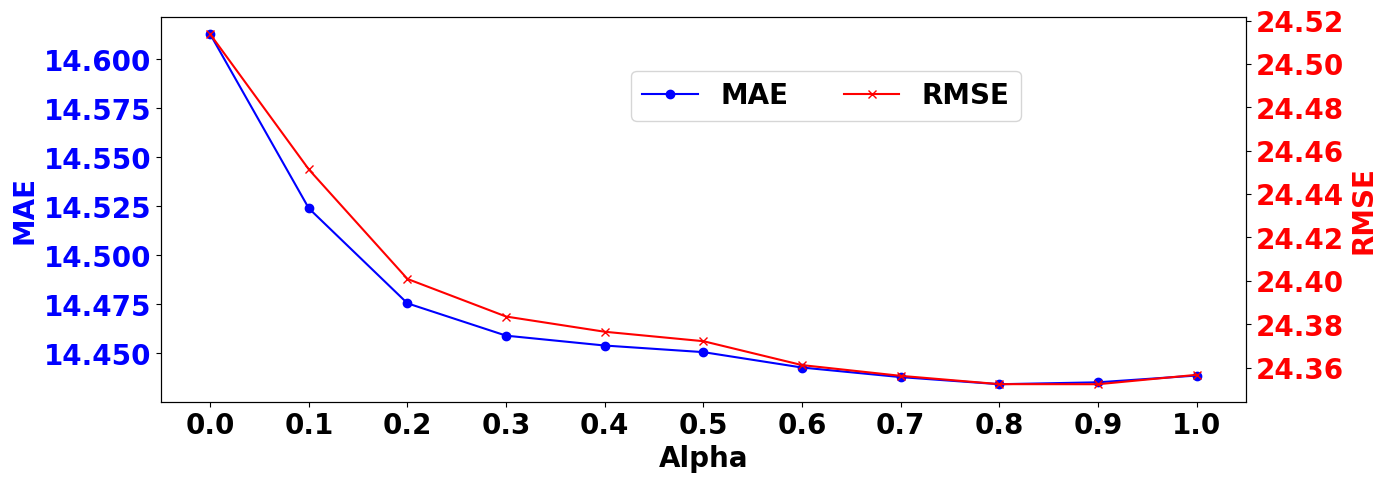}
        \caption{MAE and RMSE}
        \label{fig:pems07_mae_rmse_alpha}
    \end{subfigure}
    \hfill
    \begin{subfigure}[b]{0.45\textwidth}
        \centering
        \includegraphics[width=\textwidth]{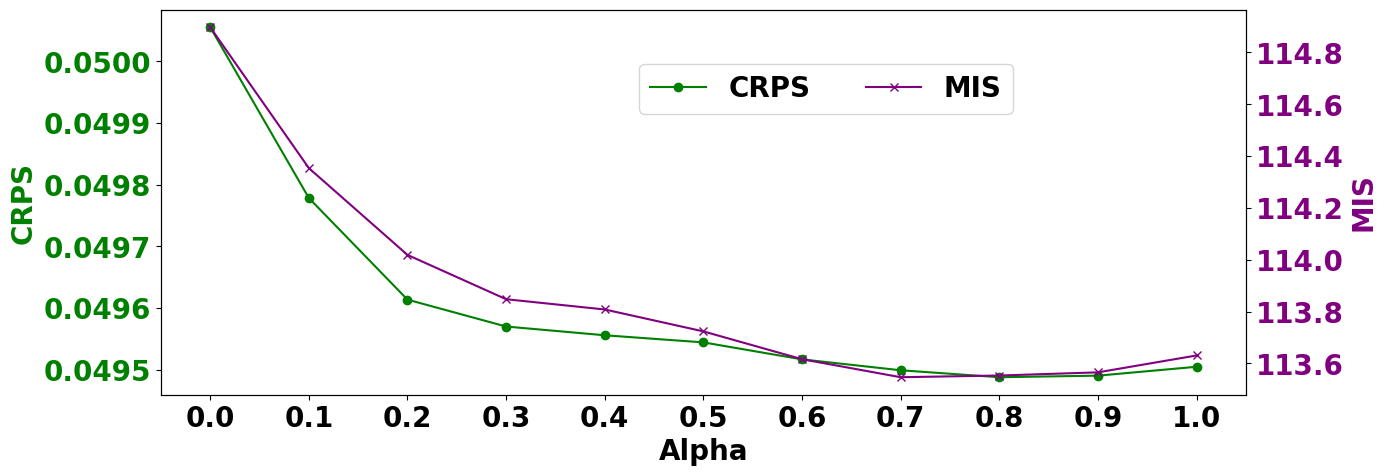}
        \caption{CRPS and MIS}
        \label{fig:pems07_crps_mis_alpha}
    \end{subfigure}
    \caption{Metrics across varying $\alpha$ in PEMS07.}
    \label{fig:alpha}
\end{figure}

\subsubsection{Full and Random Test Runs Performance Evaluation.}
The results of the full test runs on the PEMS03 and PEMS08 datasets, summarized in Table \ref{table:det-metrics}, show that ProGen outperforms other state-of-the-art deterministic models by non-trivial margins. Comparisons with probabilistic models on the full PEMS08 dataset in Table \ref{table:prob-metrics} indicate that ProGen excels in terms of all metrics as well.
Particularly, ProGen significantly outperforms DiffSTG, the discrete diffusion approach, by a large margin.
The random runs on all datasets are also conducted to further evaluate ProGen's performance against selected baselines. Deterministic models were chosen based on their performance in the full test runs of PEMS03 and PEMS08, while probabilistic models were selected for their popularity and reproducibility. The results, summarized in Table \ref{table:sup_metrics}, show ProGen's robustness and effectiveness in handling spatiotemporal traffic flow data, outperforming the selected models across most datasets and metrics. Notably, ProGen slightly underperforms in PEMS07 MAE compared to AGCRN and in PEMS08 RMSE compared to DSTAGNN. The statistical significance tests in \textcolor{black}{Appendix B} confirm ProGen's superiority over DSTAGNN, STGNRDE, and AGCRN in PEMS03 and PEMS04. While AGCRN has a marginally better MAE for PEMS07 and DSTAGNN a better RMSE for PEMS08, these differences are not statistically significant. Overall, the random test runs reinforce ProGen's superiority across datasets and metrics, establishing it as a robust and effective model for traffic flow forecasting.
Additionally, we visualized ProGen's predictions for busy nodes in PEMS03 and PEMS07 in Figure \ref{fig:pems03_04_07_08_nodes_vis}. ProGen consistently shows a narrower prediction interval than CSDI, with mean predictions closely aligning with actual data. Notably, in Figure~\ref{fig:node537_pems07}, ProGen successfully mirrors the ground truth trends, capturing patterns that other models miss, especially within the orange dashed boxes. Additional visualization is in \textcolor{black}{Appendix G3}.



\subsubsection{Reverse Iterations.}
Figure~\ref{fig:pems04_3d_waves_plot} (left) showcases the mean predictions of generated samples over 1000 iterations, approximating the continuous reverse diffusion process. Initially, brighter colors indicate significant differences between predictions and truths, but as iterations progress, the colors converge to near zero, demonstrating improved alignment with the actual values. This trend of convergence is emphasized in Figure~\ref{fig:pems04_3d_waves_plot} (right), illustrating the last timestep where most discrepancies have diminished. Only a few spikes at this final stage indicate minor remaining differences, confirming ProGen's effectiveness in refining predictions to closely match actual data across iterations. Additional visualization is in the \textcolor{black}{Appendix G1}.

\subsubsection{Prediction Distribution Analysis.}
Figure~\ref{fig:distribution_change_iteration} displays the distribution of prediction samples across iterations, demonstrating the model's progressive convergence towards actual labels. Initially marked by high variability, the predictions gradually stabilize, underscoring the model's capability to enhance forecast accuracy through iterative refinement.

\section{Ablation, Sensitivity, and Additional studies}


\subsubsection{Spatiotemporal Layer Training Comparison.}
We compared our tailored spatiotemporal layer (ST Layer) against the adaptive layer by \citet{bai2020} and the S4 layer \cite{tang2023}, both noted for their forecasting capabilities. Figure~\ref{fig:train_loss} displays the training loss curves over 100 epochs. Our ST Layer shows faster convergence and maintains a lower, more stable training loss, underscoring its effectiveness in spatiotemporal data modeling.

\subsubsection{Sensitivity to Number of Generated Samples and the Parameter \( \alpha \) in ST SDE.}
ProGen's performance, as shown in Figure~\ref{fig:sample_size}, improves with increasing sample sizes, achieving significant reductions in MAE and RMSE up to 350 samples, beyond which gains plateau, indicating an optimal sample size. Similarly, Figure~\ref{fig:alpha} reveals that adjusting \( \alpha \) in the ST SDE enhances performance up to a threshold as well, after which it declines. These findings highlight ProGen's great potential for further improvement with optimized sample sizes and \( \alpha \) values.

\subsubsection{Comparison Between Tailored ST SDE with Adaptive Selcetion and Sole Use of sub-VP SDE.}
Figure~\ref{fig:mae_stsde_subvpsde} shows the ST SDE with the adaptive reverse process delivering consistently lower MAE from the outset, confirming its superior early performance and effectiveness over iterations. Figure~\ref{fig:crps_stsde_subvpsde} further demonstrates that
although ST SDE exhibits high volatility initially, it consistently reduces CRPS compared to the sub-VP SDE.
Remarkably, while improvements in deterministic performance typically compromise probabilistic accuracy, our adaptive process enhances both, showcasing robust error management in diverse forecasting scenarios. Additional comparison is in \textcolor{black}{Appendix G2}.

\section{Conclusion}
We propose ProGen, a novel generative framework that leverages a tailored SDE and novel diffusion models to forecast spatiotemporal data. ProGen combines the strengths of deterministic and probabilistic forecasting, enabling accurate and reliable predictions for complex spatiotemporal data. Our model demonstrates superior performance compared to existing state-of-the-art deterministic and probabilistic models that are among the best in the field. Future work will focus on further improving ProGen's inference efficiency and exploring its applications in other domains where probabilistic forecasting is crucial.

\bibliography{ProGen}
\end{document}


\maketitle
\vspace{-10em}
\begin{appendices}

    \section{Proof of the convergence of our tailored spatio-temporal stochastic differential equation}

    The sub-VP SDE \citep{song2021} is:
    \begin{equation*}
        \begin{aligned}
            d\mathbf{X} & = -\frac{1}{2} \beta(t)\left( \mathbf{X} \right)dt  + \sqrt{\beta(t)(1 - e^{-2 \int_0^t \beta(s) \, ds})} \, dw,
        \end{aligned}
        \label{eq:subvp_sde}
    \end{equation*}

    The tailored Spatiotemporal SDE (ST SDE) is:
    \begin{equation*}
        \begin{aligned}
            d\mathbf{X} & = -\frac{1}{2} \beta(t)\left( \mathbf{X} - \alpha A \mathbf{X} \right)dt  + \sqrt{\beta(t)(1 - e^{-2 \int_0^t \beta(s) \, ds})} \, dw,
        \end{aligned}
        \label{eq:spatiotemporal_sde}
    \end{equation*}

    The diffusion term is:
    \[
        \sqrt{\beta(t)(1 - e^{-2 \int_0^t \beta(s) \, ds})}
    \]

    Using Equation 5.51 by \cite{sarkka2019},
    we have  \(\mathbf{V}(t) = \text{Var}(\mathbf{x}(t))\):
    \begin{equation*}
        \begin{aligned}
            \frac{d}{dt} \mathbf{V}(t) & = \beta(t) \left((I - \alpha A) \mathbf{V}(t) + \mathbf{V}(t) (I - \alpha A)^T\right) + \beta(t)(1 - e^{-2 \int_0^t \beta(s) \, ds})
        \end{aligned}
    \end{equation*}

    For convergence, we look for a steady-state solution \(\mathbf{V}_\infty\) such that \(\frac{d}{dt} \mathbf{V}(t) = 0\):

    \[
        \begin{aligned}
            0 = & \, \beta(t) \left((I - \alpha A) \mathbf{V}_\infty + \mathbf{V}_\infty (I - \alpha A)^T\right) + \beta(t)(1 - e^{-2 \int_0^t \beta(s) \, ds})
        \end{aligned}
    \]

    This simplifies to:
    \[
        (I - \alpha A) \mathbf{V}_\infty + \mathbf{V}_\infty (I - \alpha A)^T = -\left(1 - e^{-2 \int_0^t \beta(s) ds}\right)I
    \]

    Given that \( \int_0^t \beta(s) ds \to \infty \) as \( t \to \infty \), the term \( e^{-2 \int_0^t \beta(s) ds} \to 0 \). Therefore, the steady-state equation becomes:
    \[
        (I - \alpha A) \mathbf{V}_\infty + \mathbf{V}_\infty (I - \alpha A)^T = -I
    \]

    This is a Lyapunov equation, and its solution \(\mathbf{V}_\infty\) will exist and be unique if the matrix \((I - \alpha A)\) is stable, meaning all eigenvalues have positive real parts. This again could be controlled by the hyperparameter $\alpha$.

    For the proposed SDE, the mean evolution incorporates the adjacency matrix \(A\) scaled by \(\alpha\). This indicates that the mean dynamics are influenced by the spatial structure of the data. For the sub-VP SDE by \cite{song2021}, the mean evolution is only scaled by \(\beta(t)\), showing a simpler temporal decay without spatial interaction.  The variance evolution for the proposed SDE is more complex, accounting for the spatial structure through the adjacency matrix \(A\). Still the sub-VP SDE variance evolution is simpler, with the variance converging based solely on the temporal component \(\beta(t)\).

    \clearpage

    \section{Statistical Significance of Test Results Compared to Baselines}

    \begin{table}[h]
        \centering
        \renewcommand{\arraystretch}{1.1} 
        \setlength{\tabcolsep}{3pt} 
        \scriptsize 
        \begin{tabular}{l|l|l|l|l|l}
            \hline
            \textbf{Dataset}         & \textbf{Metric}       & \textbf{Model}           & \textbf{Test}             & \textbf{Statistic} & \textbf{p-value} \\ \hline
            \multirow{12}{*}{PEMS03} & \multirow{6}{*}{MAE}  & \multirow{2}{*}{DSTAGNN} & Paired t-test             & 3.90               & \textbf{0.004}   \\
                                     &                       &                          & Wilcoxon signed-rank test & 3.00               & \textbf{0.010}   \\\cline{3-6}
                                     &                       & \multirow{2}{*}{STGNRDE} & Paired t-test             & 8.28               & \textbf{0.000}   \\
                                     &                       &                          & Wilcoxon signed-rank test & 0.00               & \textbf{0.002}   \\ \cline{3-6}
                                     &                       & \multirow{2}{*}{AGCRN}   & Paired t-test             & 3.56               & \textbf{0.006}   \\
                                     &                       &                          & Wilcoxon signed-rank test & 2.00               & \textbf{0.006}   \\ \cline{2-6}
                                     & \multirow{6}{*}{RMSE} & \multirow{2}{*}{DSTAGNN} & Paired t-test             & 4.47               & \textbf{0.002}   \\
                                     &                       &                          & Wilcoxon signed-rank test & 0.00               & \textbf{0.002}   \\\cline{3-6}
                                     &                       & \multirow{2}{*}{STGNRDE} & Paired t-test             & 7.34               & \textbf{0.000}   \\
                                     &                       &                          & Wilcoxon signed-rank test & 0.00               & \textbf{0.002}   \\ \cline{3-6}
                                     &                       & \multirow{2}{*}{AGCRN}   & Paired t-test             & 3.34               & \textbf{0.009}   \\
                                     &                       &                          & Wilcoxon signed-rank test & 2.00               & \textbf{0.006}   \\ \cline{1-6}
            \multirow{12}{*}{PEMS04} & \multirow{6}{*}{MAE}  & \multirow{2}{*}{DSTAGNN} & Paired t-test             & 2.54               & \textbf{0.032}   \\
                                     &                       &                          & Wilcoxon signed-rank test & 8.00               & \textbf{0.049}   \\\cline{3-6}
                                     &                       & \multirow{2}{*}{STGNRDE} & Paired t-test             & 15.25              & \textbf{0.000}   \\
                                     &                       &                          & Wilcoxon signed-rank test & 0.00               & \textbf{0.002}   \\\cline{3-6}
                                     &                       & \multirow{2}{*}{AGCRN}   & Paired t-test             & 3.26               & \textbf{0.010}   \\
                                     &                       &                          & Wilcoxon signed-rank test & 6.00               & \textbf{0.027}   \\ \cline{2-6}
                                     & \multirow{6}{*}{RMSE} & \multirow{2}{*}{DSTAGNN} & Paired t-test             & 2.85               & \textbf{0.019}   \\
                                     &                       &                          & Wilcoxon signed-rank test & 7.0                & \textbf{0.037}   \\\cline{3-6}
                                     &                       & \multirow{2}{*}{STGNRDE} & Paired t-test             & 16.02              & \textbf{0.000}   \\
                                     &                       &                          & Wilcoxon signed-rank test & 0.00               & \textbf{0.002}   \\ \cline{3-6}
                                     &                       & \multirow{2}{*}{AGCRN}   & Paired t-test             & 4.27               & \textbf{0.002}   \\
                                     &                       &                          & Wilcoxon signed-rank test & 1.0                & \textbf{0.004}   \\ \cline{1-6}
            \multirow{12}{*}{PEMS07} & \multirow{6}{*}{MAE}  & \multirow{2}{*}{DSTAGNN} & Paired t-test             & 3.13               & \textbf{0.012}   \\
                                     &                       &                          & Wilcoxon signed-rank test & 6.00               & \textbf{0.027}   \\\cline{3-6}
                                     &                       & \multirow{2}{*}{STGNRDE} & Paired t-test             & 22.36              & \textbf{0.000}   \\
                                     &                       &                          & Wilcoxon signed-rank test & 0.00               & \textbf{0.002}   \\\cline{3-6}
                                     &                       & \multirow{2}{*}{AGCRN}   & Paired t-test             & -0.44              & 0.671            \\
                                     &                       &                          & Wilcoxon signed-rank test & 24.00              & 0.770            \\ \cline{2-6}
                                     & \multirow{6}{*}{RMSE} & \multirow{2}{*}{DSTAGNN} & Paired t-test             & 4.38               & \textbf{0.002}   \\
                                     &                       &                          & Wilcoxon signed-rank test & 2.00               & \textbf{0.006}   \\\cline{3-6}
                                     &                       & \multirow{2}{*}{STGNRDE} & Paired t-test             & 18.02              & \textbf{0.000}   \\
                                     &                       &                          & Wilcoxon signed-rank test & 0.00               & \textbf{0.002}   \\ \cline{3-6}
                                     &                       & \multirow{2}{*}{AGCRN}   & Paired t-test             & 0.85               & 0.418            \\
                                     &                       &                          & Wilcoxon signed-rank test & 17.00              & 0.322            \\ \cline{1-6}
            \multirow{12}{*}{PEMS08} & \multirow{6}{*}{MAE}  & \multirow{2}{*}{DSTAGNN} & Paired t-test             & 0.56               & 0.592            \\
                                     &                       &                          & Wilcoxon signed-rank test & 26.00              & 0.922            \\\cline{3-6}
                                     &                       & \multirow{2}{*}{STGNRDE} & Paired t-test             & 8.81               & \textbf{0.000}   \\
                                     &                       &                          & Wilcoxon signed-rank test & 0.00               & \textbf{0.002}   \\ \cline{3-6}
                                     &                       & \multirow{2}{*}{AGCRN}   & Paired t-test             & 1.53               & 0.159            \\
                                     &                       &                          & Wilcoxon signed-rank test & 14.00              & 0.193            \\ \cline{2-6}
                                     & \multirow{6}{*}{RMSE} & \multirow{2}{*}{DSTAGNN} & Paired t-test             & -0.46              & 0.653            \\
                                     &                       &                          & Wilcoxon signed-rank test & 24.00              & 0.770            \\\cline{3-6}
                                     &                       & \multirow{2}{*}{STGNRDE} & Paired t-test             & 5.99               & \textbf{0.000}   \\
                                     &                       &                          & Wilcoxon signed-rank test & 0.00               & \textbf{0.002}   \\ \cline{3-6}
                                     &                       & \multirow{2}{*}{AGCRN}   & Paired t-test             & 1.28               & 0.231            \\
                                     &                       &                          & Wilcoxon signed-rank test & 16.00              & 0.275            \\ \cline{1-6}
        \end{tabular}
        \caption{Paired t-test and Wilcoxon signed-rank test results for MAE and RMSE comparing AGCRN, DSTAGNN, STGNRDE models with ProGen on PEMS03, PEMS04, PEMS07 and PEMS08 datasets.}
        \label{tab:stats_results}
    \end{table}

    The statistical significance of the test results over well-performed deterministic baselines is shown in Table \ref{tab:stats_results}. The p-values are calculated using the paired t-test and Wilcoxon signed-rank test. The p-values are shown in bold if they are less than 0.05, indicating that the test results are statistically significant, meaning the observed differences are not due to random chance. We can safely count out PEMS03 and PEMS08 since we have conducted full tests on these datasets. For PEMS04 and PEMS07, the test results are statistically significant except for AGCRN in PEMS07. Our random tests in the main paper show that AGCRN slightly outperforms ProGen in PEMS07, and the statistical test results imply that the difference is not statistically significant. Overall, the test results show that ProGen outperforms the well-performed deterministic baselines, and the performance of ProGen over other state-of-arts are statistically significant.

    \clearpage
    \section{Data Description}

    \begin{table}[h]
        \centering
        \begin{adjustbox}{max width=\columnwidth}
            \begin{tabular}{c|c|c|c|c|c}
                \hline
                Datasets & \# Nodes & \# Edges & \# Samples & Time Span       & Frequency \\
                \hline
                PEMS03   & 358      & 547      & 26208      & 09/2018-11/2018 & 5 min     \\
                PEMS04   & 307      & 340      & 16992      & 01/2018-02/2018 & 5 min     \\
                PEMS07   & 883      & 866      & 28224      & 05/2017-08/2017 & 5 min     \\
                PEMS08   & 170      & 295      & 17856      & 07/2016-08/2016 & 5 min     \\
                \hline
            \end{tabular}
        \end{adjustbox}
        \caption{Summary Statistics of PEMS Datasets.}
        \label{table:dataset_stats}
    \end{table}

    \section{Evaluation Metrics}

    \begin{itemize}
        \item \textbf{MAE}:
              \[
                  \text{MAE} = \frac{1}{n} \sum_{i=1}^{n} \left| y_i - \hat{y}_i \right|
              \]
              where \( y_i \) is the actual value and \( \hat{y}_i \) is the predicted value.

        \item \textbf{RMSE}:
              \[
                  \text{RMSE} = \sqrt{\frac{1}{n} \sum_{i=1}^{n} (y_i - \hat{y}_i)^2}
              \]
              where \( y_i \) is the actual value and \( \hat{y}_i \) is the predicted value.

        \item \textbf{CRPS} (Continuous Ranked Probability Score):
              \[
                  \text{CRPS}(F, y) = \int_{-\infty}^{\infty} \left( F(x) - \mathbf{1}\{x \geq y\} \right)^2 dx
              \]
              where \( F \) is the cumulative distribution function of the forecast and \( y \) is the observed value. CRPS evaluates the accuracy of the predicted cumulative distribution function compared to the actual value. It generalizes the concept of MAE to probabilistic forecasts, measuring both the reliability and sharpness of the predicted distributions.

        \item \textbf{MIS} (Mean Interval Score):
              \begin{equation*}
                  \begin{aligned}
                      \text{MIS} = \frac{1}{n} \sum_{i=1}^{n} \Bigg( (U_i - L_i) + \frac{2}{\alpha} (L_i - y_i) \mathbf{1}\{y_i < L_i\}
                      + \frac{2}{\alpha} (y_i - U_i) \mathbf{1}\{y_i > U_i\} \Bigg)
                  \end{aligned}
              \end{equation*}
              where \( [L_i, U_i] \) is the predicted interval for the \( i \)-th observation, \( y_i \) is the actual value, and \( \alpha \) is the nominal coverage probability. MIS assesses the quality of predictive intervals, balancing interval width $(U_i - L_i)$ with the penalty for observations falling outside the predicted interval. It provides insight into both the calibration and sharpness of the prediction intervals.
    \end{itemize}

    \clearpage
    \section{Algorithms for Adaptive Reverse Sampling Prediction Process and Prediction Details}

    \begin{algorithm}[htbp!]
        \caption{An Overview of the Adaptive Reverse Sampling Prediction Process}
        \label{alg:reverse_workflow}
        \textbf{Input}: Trained ProGen model $s_{\theta^*}$, conditional inputs $\mathcal{C}$ (including graph structure $\mathcal{G}$, historical data $\mathbf{X_H}$, positional indicators $\mathbf{P_H}$) \\
        \textbf{Parameter}: Number of samples $N$, number of discretized time steps $K$ in the reverse process\\
        \textbf{Output}: Predicted future states $\mathbf{X^0_F}$, optimal predictions $\mathbf{X^*_F}$ during the reverse process
        \begin{algorithmic}[1]
            \STATE Initialize $N$ samples $\mathbf{\tilde{X}^1_\mathbf{F}} \sim \mathcal{N}(\mathbf{0}, \mathbf{I})$
            \STATE Set best selected metric to infinity and initialize best predictions as None.
            \FOR{$k = K-1$ \textbf{downto} $0$}
            \STATE Set \( t = \frac{k}{K} \) to convert the loop index to normalized time
            \FOR{each sample $n = 1$ \textbf{to} $N$}
            \STATE Perform reverse sampling for sample $n$ using the ST SDE and sub-VP SDE described in Algorithm~\ref{alg:prediction_process} in Appendix E to obtain $\mathbf{\tilde{X}^{t-\frac{1}{K}}_{\mathbf{F}, n}}$:
            \[
                \mathbf{\tilde{X}}^{t-\frac{1}{K}}_{\mathbf{F}, n} = \mathcal{F}(\text{Selected SDE}, \mathbf{\tilde{X}}^t_{\mathbf{F}, n}, t, \mathcal{C})
            \]
            \ENDFOR

            \STATE Compute evaluation metrics for both SDE equations for sample $n$.
            \IF{Metric for ST-SDE is better}
            \STATE Store predictions $\mathbf{\tilde{X}}^{t-\frac{1}{K}}_{\mathbf{F}, n}$ and metrics using ST SDE for sample $n$.
            \ELSE
            \STATE Store predictions $\mathbf{\tilde{X}}^{t-\frac{1}{K}}_{\mathbf{F}, n}$ and metrics using sub-VP SDE for sample $n$.
            \ENDIF
            \IF{Current metric for sample $n$ is better than the best metric}
            \STATE Update the best predictions $\mathbf{\tilde{X}^*_{\mathbf{F}, n}}$.
            \ENDIF
            \ENDFOR
            \STATE \textbf{return} Predicted future states $\mathbf{X^0_F}$ and optimal predictions $\mathbf{\tilde{X}^*_F}$.
        \end{algorithmic}
    \end{algorithm}

    The Adaptive Reverse Sampling Prediction in Algorithm~\ref{alg:reverse_workflow} employs the ProGen model for generating predictions by reversing the diffusion process. Starting with $N$ sampels of data points from a Gaussian distribution, the algorithm iteratively denoises these points across \( K \) discretized time steps. Each step involves comparing two stochastic differential equation (SDE) models: the Spatiotemporal SDE (ST-SDE) and a subordinate VP SDE (sub-VP SDE). The algorithm selects the predictions from the SDE equation that performs better based on specific evaluation metrics (MAE in our experiments) at each diffusion time step, updating the best predictions at the same time. This adaptive approach allows the algorithm to optimize prediction accuracy by dynamically choosing the more effective SDE model for each reverse sampling step, thereby refining the predictions towards the original data distribution.

    \begin{algorithm}[htbp!]
        \caption{Reverse Sampling Process for Prediction with Tailored ST-SDE and sub-VP SDE}
        \label{alg:prediction_process}
        \textbf{Input}: Current sample $\tilde{\mathbf{X^t_F}}$, discretized timestep index $k$, total discretized timesteps $K$, graph structure $\mathcal{G}$, historical data $\mathbf{X_H}$, positions of week and day information $\mathbf{P_H}$ \\
        \textbf{Parameter}: Score function $s_{\theta^*}$, use flag for ST SDE $f$, time coefficients $\beta_0$ and $\beta_1$ \\
        \textbf{Output}: Prediction samples $\mathbf{\tilde{X}^{t-\frac{1}{K}}_F}$ at next diffusion timestep.
        \begin{algorithmic}[1]
            \STATE Compute normalized time $t \leftarrow \frac{k}{K}$.
            \STATE Neighboring mean effect $\mathbf{M}$ by performing Einstein summation on $A$ and $\tilde{\mathbf{X^t_F}}$.
            \STATE Time-dependent coefficient
            $\beta_t\leftarrow \beta_0 + t \cdot (\beta_1 - \beta_0)$
            \STATE Diffusion term: $\mathbf{G} \leftarrow \sqrt{\beta_t \cdot \left(1 - e^{-2 \beta_0 t - (\beta_1 - \beta_0) t^2}\right)}$
            \IF{$f$ is TRUE}
            \STATE
            $\text{Dift Term }{D} \leftarrow
                -\frac{1}{2}
                \cdot \beta_t
                \cdot (\tilde{\mathbf{X}}^t_\mathbf{F} - \alpha  \mathbf{M})
            $
            \ELSE
            \STATE
            $\text{Dift Term }{D} \leftarrow
                -\frac{1}{2}
                \cdot \beta_t
                \cdot \tilde{\mathbf{X}}^t_\mathbf{F}
            $
            \ENDIF
            \STATE Score $s \leftarrow s_{{\theta}^{*}}(\tilde{\mathbf{X}}^t_\mathbf{F}, \mathbf{X_H}, \mathcal{G}, t, \mathbf{P_H})$.
            \STATE Same as the first part of Equation~2 in the main paper, update $D$ for reverse sampling $D_{rev} \leftarrow D - G^2s$
            \STATE The mean of the updated sample $[\tilde{\mathbf{X}}^t_\mathbf{F}]_{mean} \leftarrow \tilde{\mathbf{X}}^t_\mathbf{F} - D_{rev}$.
            \STATE Random noise $z \sim \mathcal{N}(\mathbf{0}, \mathbf{I})$
            \STATE Move the sample $\tilde{\mathbf{X}}^t_\mathbf{F} \leftarrow [\tilde{\mathbf{X}}^t_\mathbf{F}]_{mean} + D * z$.
            \STATE \textbf{return} Prediction samples $\mathbf{\tilde{X}}^{t-\frac{1}{K}}_\mathbf{F}$
        \end{algorithmic}
    \end{algorithm}

    The Reverse Sampling Process for Prediction with Tailored ST-SDE and sub-VP SDE in Algorithm~\ref{alg:prediction_process} employs selected SDE equation to model the reverse diffusion process for prediction. Given a current sample $\tilde{\mathbf{X^t_F}}$ at diffusion timestep $t = \frac{k}{K}$, where $k$ is the discretized timestep index and $K$ is the total number of discretized timesteps, the algorithm advances by calculating the neighboring mean effect through Einstein summation on the adjacency matrix and the sample. Depending on whether ST SDE is enabled (flagged by $f$), a corresponding drift term is chosen and adjusted dynamically alongside a diffusion term that evolves based on temporal coefficients $\beta_0$ and $\beta_1$. The sample is further refined by a score function $s_{\theta^*}$, taking into account historical data, the graph structure, and other contextual details. This score adjusts the drift term for effective reverse sampling, as detailed in Equation~2 from the main paper. The modified sample is then updated by incorporating random noise, and the result, $\mathbf{\tilde{X}^{t-\frac{1}{K}}_F}$, represents a sample of the prediction at the next diffusion timestep.

    \clearpage

    \section{Baselines and Experiments Setup}
    We compare ProGen against a broad range of traditional and state-of-the-art baseline models to establish its efficacy in both deterministic and probabilistic forecasting contexts.

    \subsection{Baselines}

    \subsubsection{Deterministic Baselines}
    The deterministic baselines include:
    \begin{itemize}
        \item \textbf{ARIMA}: A traditional time series forecasting model.
        \item \textbf{FCLSTM} \citep{sutskever2014}: Utilizes multilayered LSTM networks for sequence-to-sequence learning.
        \item \textbf{MTGNN} \citep{wu2020}: A framework leveraging Graph Neural Networks (GNNs) for multivariate time series forecasting that automatically extracts variable relations and captures spatial and temporal dependencies.
        \item \textbf{ASTGCN} \citep{zhu2020}: An attribute-augmented spatiotemporal graph convolutional network designed for traffic forecasting that integrates both dynamic and static external factors.
        \item \textbf{AGCRN} \citep{bai2020}: Employs adaptive modules for node-specific pattern learning and automatic graph generation.
        \item \textbf{STSGCN} \citep{song2020a}: A spatial-temporal network that captures complex correlations and heterogeneities.
        \item \textbf{DSTAGNN} \citep{lan2022}: Utilizes dynamic spatial-temporal aware graphs and an enhanced graph neural network architecture to capture complex dependencies.
        \item \textbf{STGNCDE} \citep{choi2022}: Combines graph convolutional networks and neural controlled differential equations for spatio-temporal processing.
        \item \textbf{STGNRDE} \citep{choi2023}: An improved version of STGNCDE that combines graph convolutional networks and neural rough differential equations for spatio-temporal processing.
    \end{itemize}

    \subsubsection{Probabilistic Baselines}
    The probabilistic baselines include:
    \begin{itemize}
        \item \textbf{Latent ODE} \citep{rubanova2019}: Uses ordinary differential equations to handle continuous-time hidden dynamics.
        \item \textbf{DeepAR} \citep{salinas2020}: Utilizes an autoregressive recurrent neural network trained on multiple time series.
        \item \textbf{CSDI} \citep{tashiro2021}: A time series imputation method using conditional score-based diffusion models.
        \item \textbf{TimeGrad} \citep{rasul2021}: An autoregressive model that uses diffusion probabilistic models.
        \item \textbf{MC Dropout} \citep{wu2021a}: Estimates uncertainty in deep learning by employing dropout during both training and inference.
        \item \textbf{DiffSTG} \citep{wen2023}: A framework integrating discrete denoising diffusion models with graph neural networks for probabilistic forecasting.
    \end{itemize}

    \subsection{Experiments Setup}

    Related code and model checkpoints for each dataset have been saved and will be made available.
    All models were trained using a system equipped with a single NVIDIA A800-SXM4-80GB GPU.
    The training duration varied across datasets, with the specific details provided below:

    \begin{itemize}
        \item \textbf{PEMS03}: The model contains approximately 19 million parameters and required around 13 hours for training. The optimal hyperparameters for this dataset are as follows: a spatial-temporal layer dimension of 64, hidden dimensions of 212, position embedding dimensions for day and week of 16, three residual blocks, channel multipliers for downsampling and upsampling of [1, 2, 3, 4], drift term $\alpha = 0.61$, batch size of 32, and a learning rate of 0.0018.

        \item \textbf{PEMS04}: The model contains approximately 35 million parameters and required around 12 hours for training. The optimal hyperparameters identified are a spatial-temporal layer dimension of 256, hidden dimensions of 242, position embedding dimensions for day and week of 128, three residual blocks, channel multipliers for downsampling and upsampling of [1, 2], drift term $\alpha = 0.61$, batch size of 32, and a learning rate of 0.0030.

        \item \textbf{PEMS07}: The model contains approximately 1.3 million parameters and required around 12 hours for training. The optimal hyperparameters include a spatial-temporal layer dimension of 32, hidden dimensions of 97, position embedding dimensions for day and week of 128, three residual blocks, channel multipliers for downsampling and upsampling of [1, 2], drift term $\alpha = 0.51$, batch size of 32, and a learning rate of 0.0029.

        \item \textbf{PEMS08}: The model contains approximately 1.6 million parameters and required around 6 hours for training. The optimal hyperparameters identified for this dataset are as follows: a spatial-temporal layer dimension of 64, hidden dimensions of 95, position embedding dimensions for day and week of 16, four residual blocks, channel multipliers for downsampling and upsampling of [1, 2], drift term $\alpha = 0.61$, batch size of 32, and a learning rate of 0.0038.
    \end{itemize}

    The training times, while substantial, are consistent with observations in other diffusion models, particularly those targeting image generation, where models often have millions of parameters \cite{ho2020, dhariwal2021, song2021}. Recent studies have continued to capture the community's attention regarding these challenges \cite{wang2023b, xu2024}. These considerations are further discussed in the conclusion, where potential future directions of this work are proposed.

    For inference, during reverse prediction, continuous time from 0 to 1 was discretized into 1000 iterations, yielding 30 samples for each dataset. We utilized 8 NVIDIA A800 GPUs to process the full test runs for the PEMS03 and PEMS08 datasets, with batch sizes of 128 for PEMS08 and 32 for PEMS03. For the PEMS04 and PEMS07 datasets, due to the large volume of data and the graph size, random run tests were employed to ensure computational efficiency and fair comparison. These tests were conducted in 11 batches, and the results were subjected to statistical significance tests to validate the fairness of ProGen's performance.

    The experiments were conducted on a Linux Ubuntu 20.04 operating system, with a critical software stack that included PyTorch running in a Python 3.9 environment with CUDA 11.7 support. Hyperparameter optimization was performed using Hyperopt \citep{bergstra2013}, an automated approach for hyperparameter tuning. The optimization was guided by the validation loss, utilizing the provided range of hyperparameters when no default settings were available in the original paper.

    \clearpage
    \section{Additional Experiments}
    \vspace{-1.5em}

    \subsection{Visualization of the reverse prediction process}
    \vspace{-1.5em}
    \begin{figure*}[h]
        \centering
        \includegraphics[width=\textwidth]{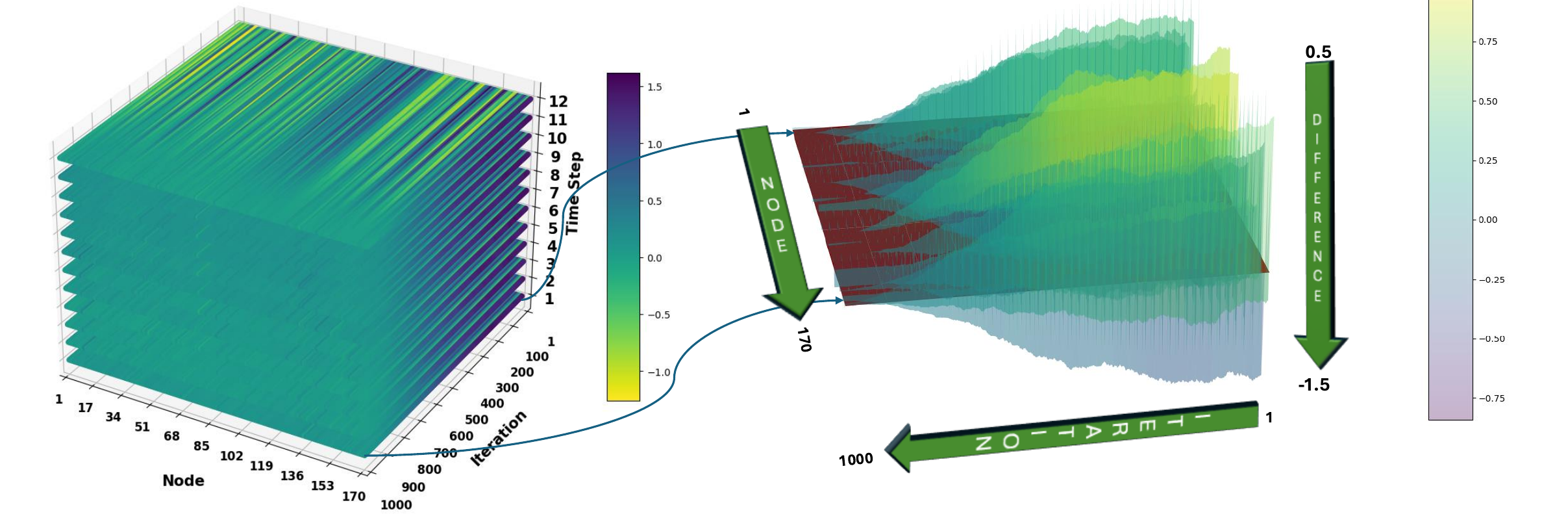}
        \caption{Visualization of the difference between the mean predictions and the truths for the average test data in PEMS08.}
        \label{fig:pems08_3d_waves_plot}
    \end{figure*}

    Figure~\ref{fig:pems08_3d_waves_plot} in this appendix provides a visualization similar to the one in the main paper, illustrating the difference between the mean predictions of generated samples and the ground truths over 1000 iterations, approximating the continuous reverse diffusion process for the PEMS08 dataset. At the beginning of the first iteration, brighter colors indicate a significant mean difference between the predictions and the ground truths. As the process progresses, the color gradually shifts to light green, signifying a difference close to zero. This indicates that the model's predictions converge towards the actual labels as the stochastic differential equations are applied at each timestep. Additionally, the visualization highlights the first timestep, where an initial spike in the difference at the first iteration gradually diminishes, nearly disappearing by the 1000th iteration, with only a few spikes remaining, demonstrating improved predictions over time.

    \vspace{-2em}
    \subsection{SDE comparison}

    \begin{figure*}[htbp!]
        \centering
        \begin{subfigure}[b]{0.48\textwidth}
            \centering
            \includegraphics[width=\textwidth]{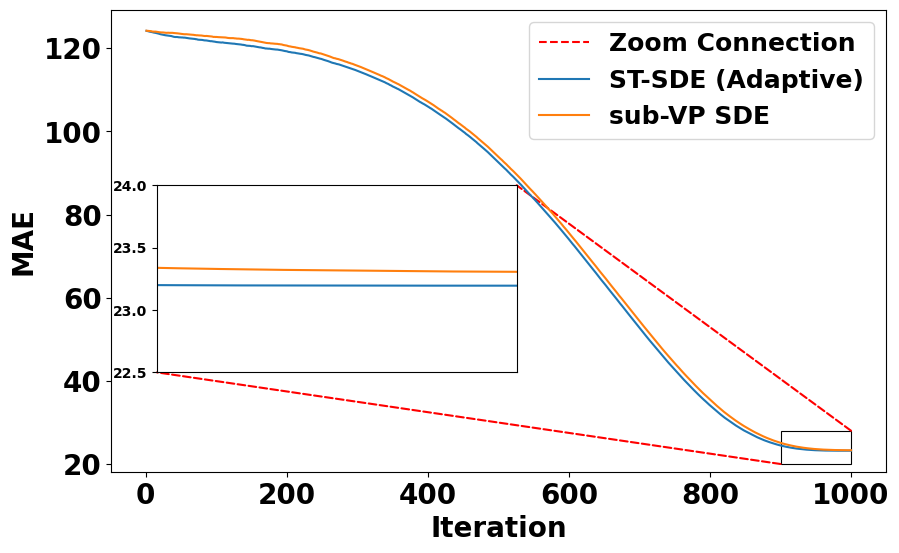}
            \caption{MAE in PEMS04}
            \label{fig:pems04_mae_stsde_subvpsde}
        \end{subfigure}
        \hfill
        \begin{subfigure}[b]{0.48\textwidth}
            \centering
            \includegraphics[width=\textwidth]{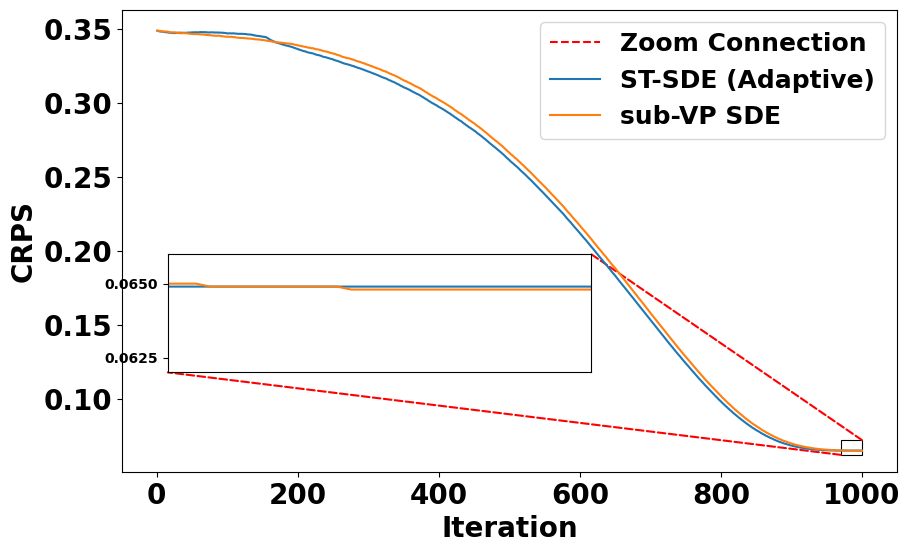}
            \caption{CRPS in PEMS04}
            \label{fig:pems04_crps_stsde_subvpsde}
        \end{subfigure}
        \hfill
        \begin{subfigure}[b]{0.48\textwidth}
            \centering
            \includegraphics[width=\textwidth]{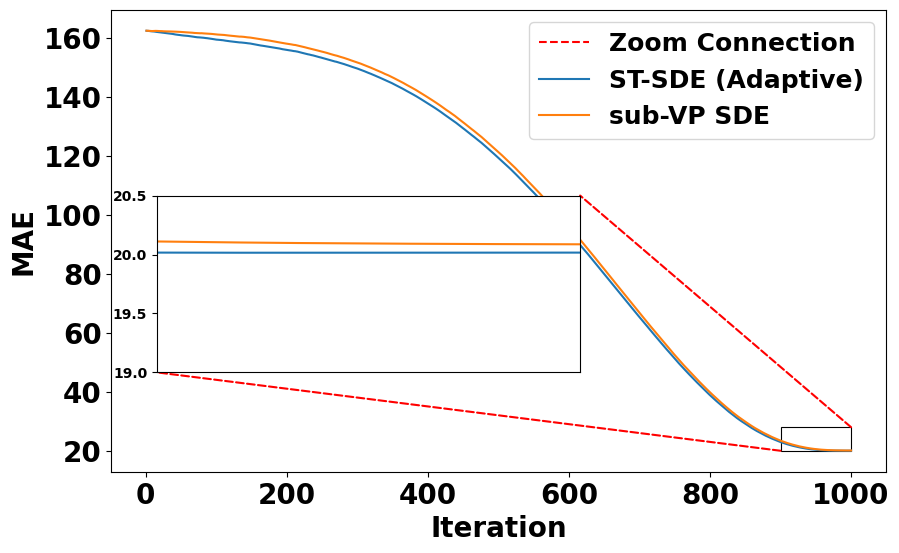}
            \caption{MAE in PEMS07}
            \label{fig:pems07_mae_stsde_subvpsde}
        \end{subfigure}
        \hfill
        \begin{subfigure}[b]{0.48\textwidth}
            \centering
            \includegraphics[width=\textwidth]{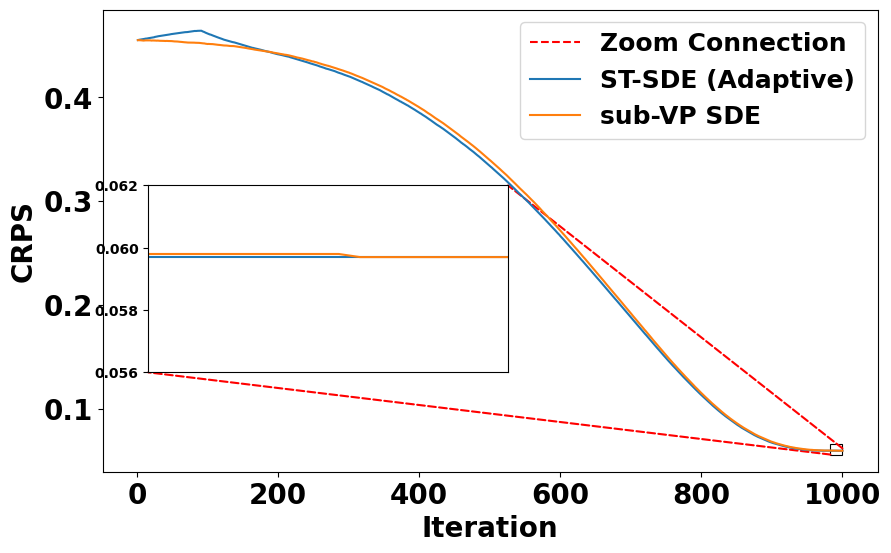}
            \caption{CRPS in PEMS07}
            \label{fig:pems07_crps_stsde_subvpsde}
        \end{subfigure}
        \caption{Performance comparison between the tailored spatial-temporal SDE and the subVP-SDE.}
        \label{fig:sde_comparison}
    \end{figure*}

    Figure~\ref{fig:sde_comparison} in the appendix presents performance comparison between the tailored spatial-temporal SDE (ST SDE) and the subVP-SDE for the PEMS04 and PEMS07 datasets. As discussed in the main paper, the MAE results for both datasets demonstrate a clear advantage of our ST SDE over the subVP-SDE during convergence, highlighting the effectiveness of the ST SDE in capturing more nuanced spatiotemporal characteristics and utilizing them during iteration.

    However, the CRPS, which is a probabilistic metric, shows less distinction between the two methods across these datasets. We attribute this to two main factors. First, there is a general expectation that improving deterministic performance may lead to a reduction in probabilistic performance due to a trade-off between specificity and diversity. Second, the tuning of the hyperparameter $\alpha$ in our ST SDE might not be optimized for probabilistic performance, as it was primarily tuned for the PEMS08 dataset. Despite this, the performance of the ST SDE in other datasets remains competitive.

    \subsection{Visualization of predictions among strong baselines}
    \begin{figure*}[h!]
        \centering
        \begin{subfigure}[b]{0.45\textwidth}
            \centering
            \includegraphics[width=\textwidth]{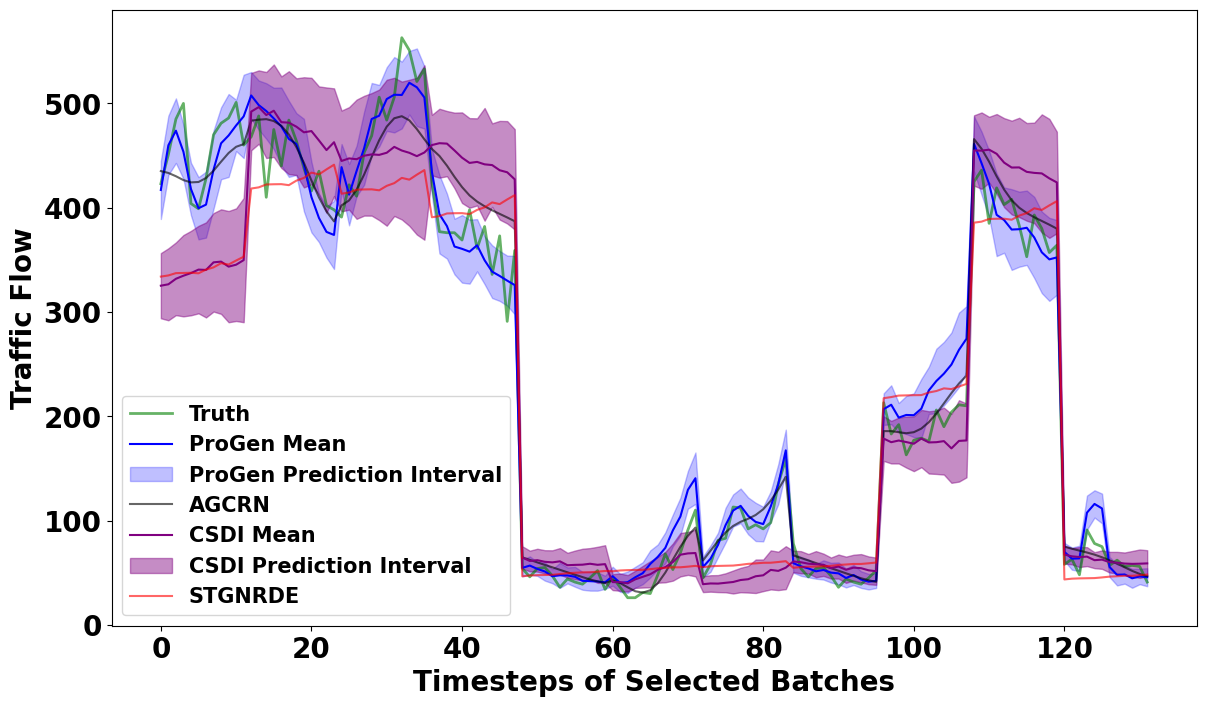}
            \caption{Node 10 in PEMS03.}
            \label{fig:node10_pems03}
        \end{subfigure}
        \hfill
        \begin{subfigure}[b]{0.45\textwidth}
            \centering
            \includegraphics[width=\textwidth]{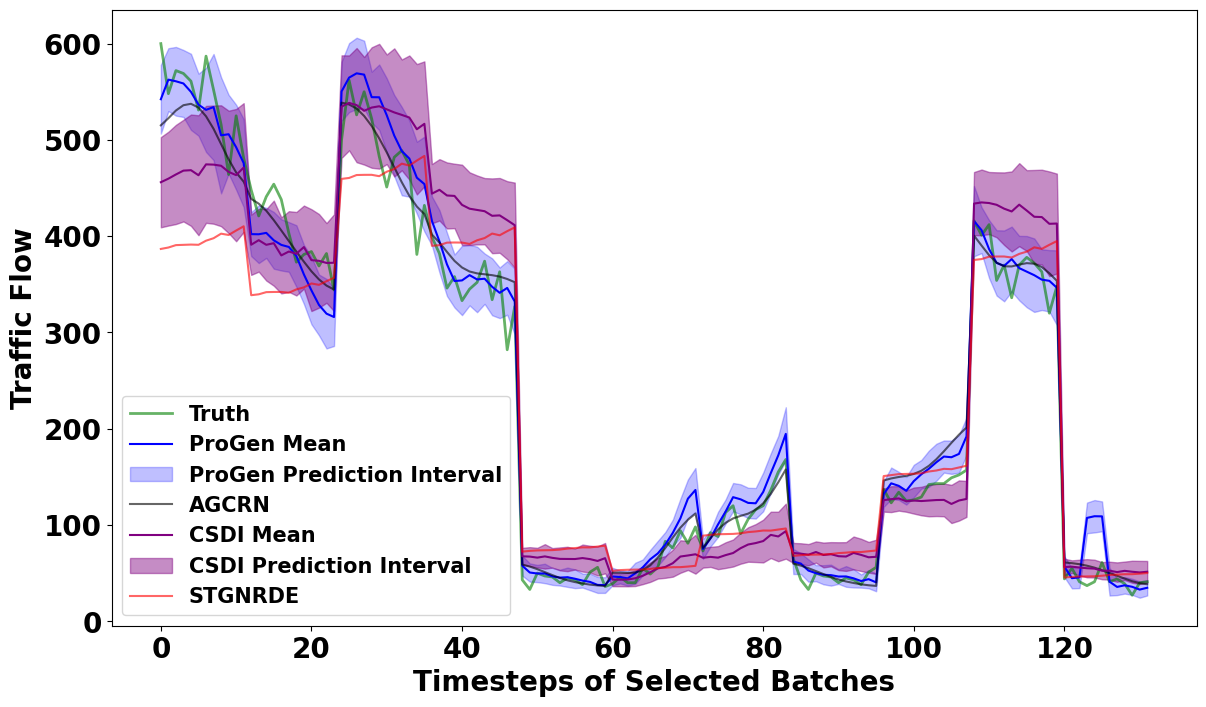}
            \caption{Node 333 in PEMS03.}
            \label{fig:node333_pems03}
        \end{subfigure}
        \hfill
        \begin{subfigure}[b]{0.45\textwidth}
            \centering
            \includegraphics[width=\textwidth]{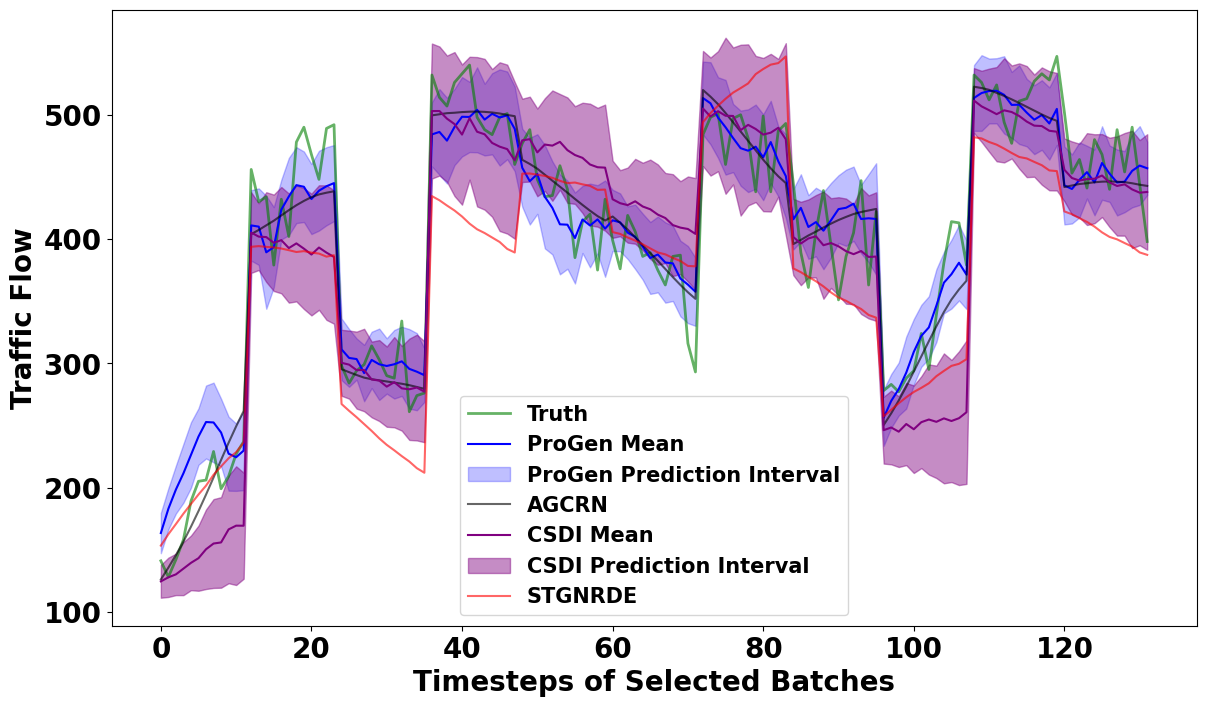}
            \caption{Node 63 in PEMS04.}
            \label{fig:node63_pems04}
        \end{subfigure}
        \hfill
        \begin{subfigure}[b]{0.45\textwidth}
            \centering
            \includegraphics[width=\textwidth]{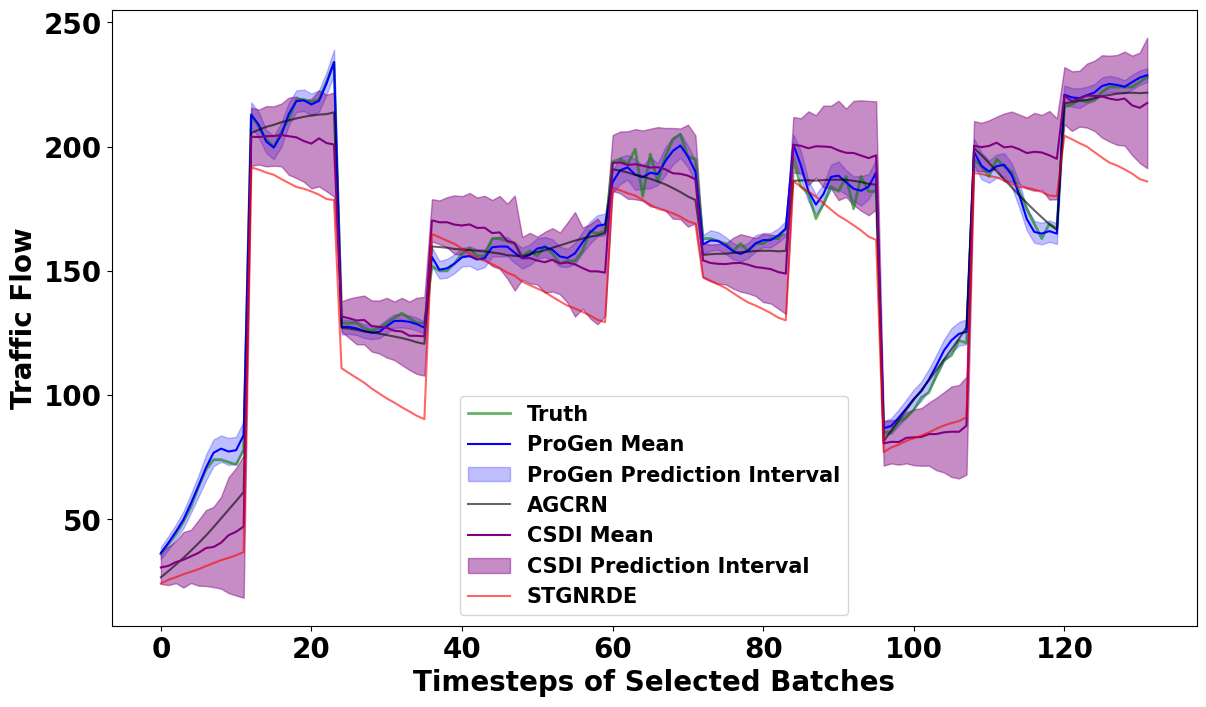}
            \caption{Node 200 in PEMS04.}
            \label{fig:node200_pems04}
        \end{subfigure}
        \hfill
        \begin{subfigure}[b]{0.45\textwidth}
            \centering
            \includegraphics[width=\textwidth]{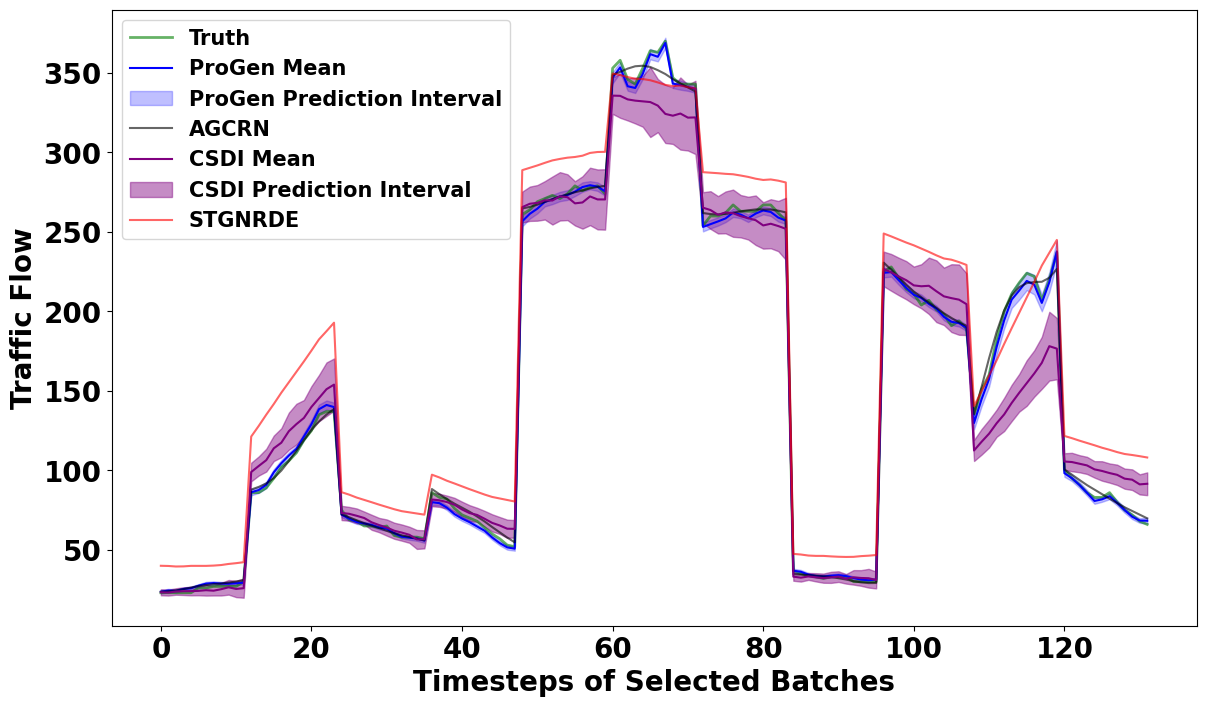}
            \caption{Node 65 in PEMS07.}
            \label{fig:node65_pems07}
        \end{subfigure}
        \hfill
        \begin{subfigure}[b]{0.45\textwidth}
            \centering
            \includegraphics[width=\textwidth]{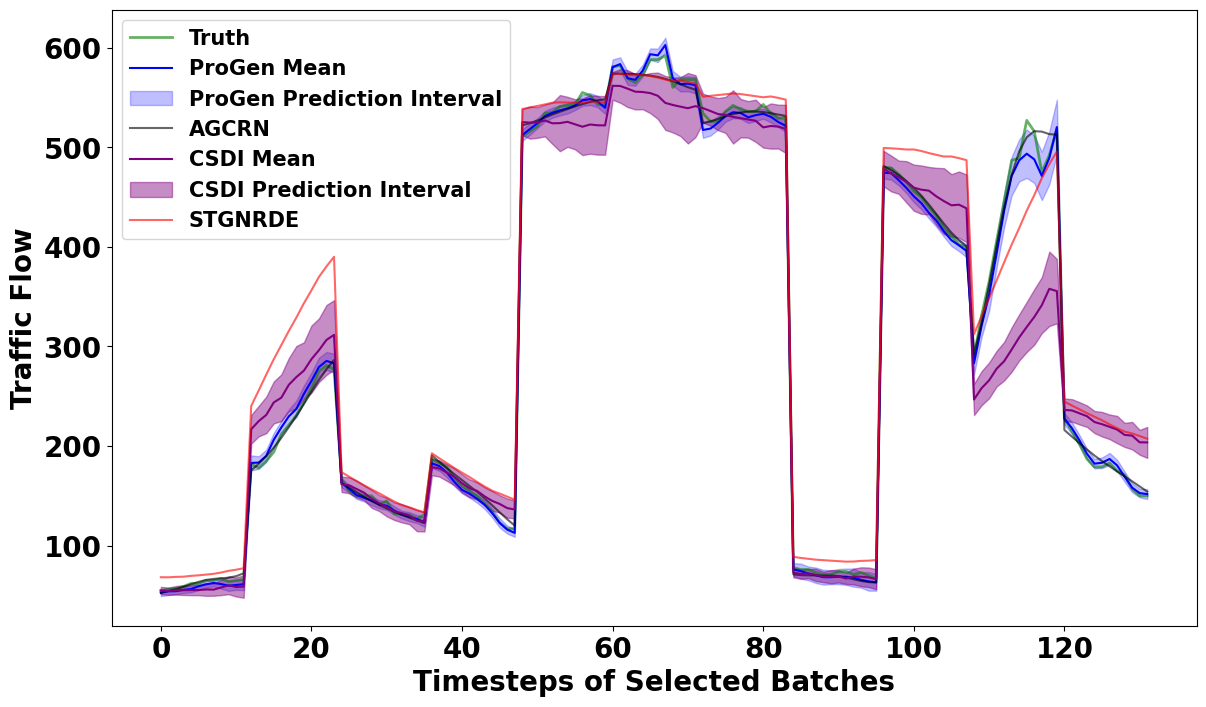}
            \caption{Node 345 in PEMS07.}
            \label{fig:node345_pems07}
        \end{subfigure}
        \hfill
        \begin{subfigure}[b]{0.45\textwidth}
            \centering
            \includegraphics[width=\textwidth]{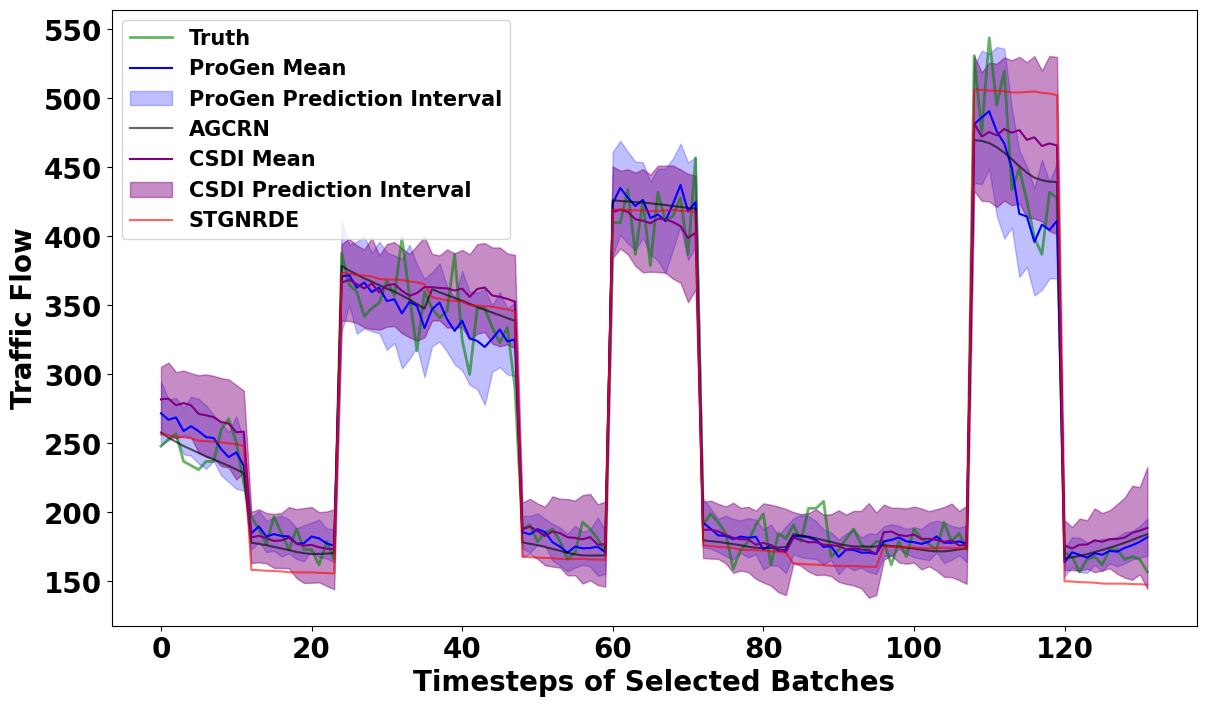}
            \caption{Node 71 in PEMS08.}
            \label{fig:node71_pems08}
        \end{subfigure}
        \hfill
        \begin{subfigure}[b]{0.45\textwidth}
            \centering
            \includegraphics[width=\textwidth]{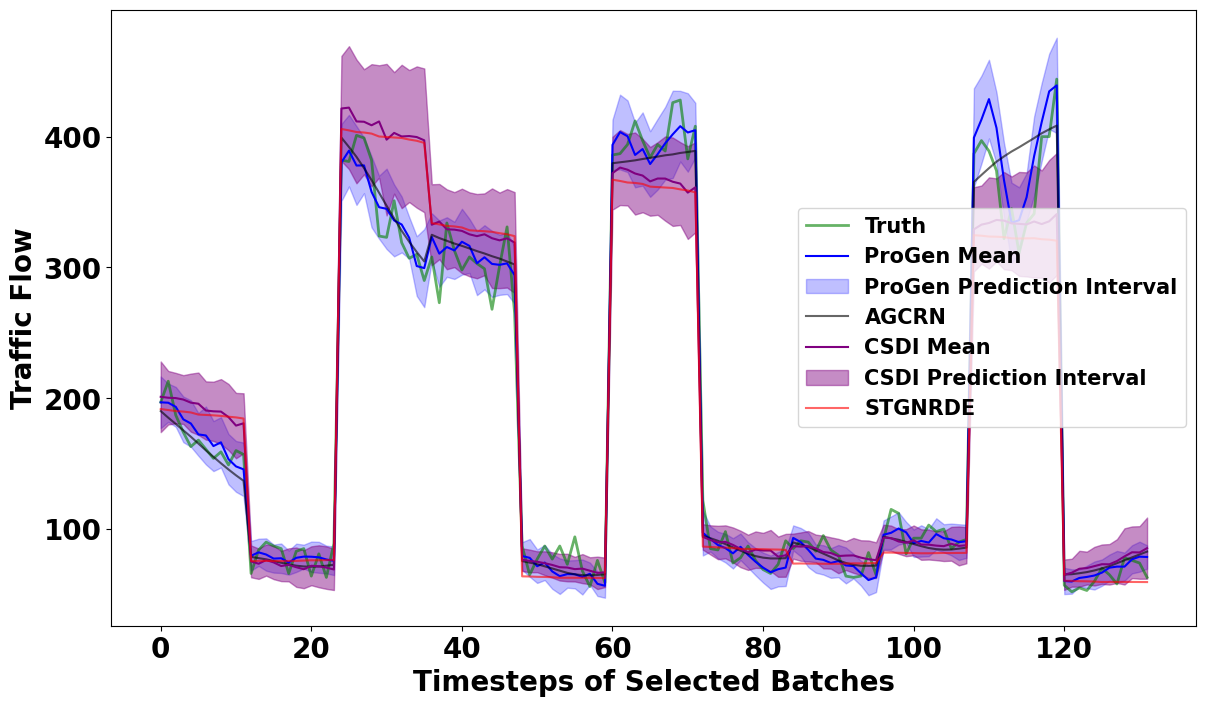}
            \caption{Node 152 in PEMS08.}
            \label{fig:node152_pems08}
        \end{subfigure}
        \caption{Visualization of predictions among strong baselines.}
        \label{fig:pems03_04_07_08_nodes_vis}
    \end{figure*}

\end{appendices}

Additional visualizations comparing ProGen with other strong deterministic and probabilistic baselines can be found in Figure~\ref{fig:pems03_04_07_08_nodes_vis} in this appendix. Across different datasets, ProGen consistently outperforms the other baselines in both deterministic and probabilistic forecasting. For instance, in Figure~\ref{fig:node10_pems03}, ProGen's mean predictions more accurately capture the trend of the ground truths around the 80-step mark. Similarly, in Figure~\ref{fig:node200_pems04}, ProGen exhibits a narrower prediction interval compared to CSDI, a well-known and high-performing model among probabilistic methods in our experiments. Even in cases where extreme values of the ground truths occur, ProGen outperforms CSDI and other deterministic models in capturing these extreme values. Overall, ProGen demonstrates narrower prediction intervals and more accurate mean predictions compared to other baselines, highlighting its superior performance in both deterministic and probabilistic forecasting.

\clearpage
\bibliographystyle{apalike}
\bibliography{ProGen}